\documentclass[10pt,twocolumn,letterpaper]{article}

\usepackage{cvpr}              

\usepackage[dvipsnames]{xcolor}

\definecolor{cvprblue}{rgb}{0.21,0.49,0.74}
\usepackage[pagebackref,breaklinks,colorlinks,citecolor=cvprblue]{hyperref}
\usepackage{bbm}
\usepackage{multirow}
\usepackage{bbding}
\usepackage{float}
\usepackage[accsupp]{axessibility} 

\def\paperID{1900} 
\def\confName{CVPR}
\def\confYear{2024}

\newtheorem{definition}{Definition}

\newtheorem{recommendation}{Recommendation}

\usepackage{amsmath}

\title{Rethinking the Evaluation Protocol of Domain Generalization}

\author{Han Yu, Xingxuan Zhang, Renzhe Xu, Jiashuo Liu, Yue He, Peng Cui\thanks{Corresponding author}\\
Department of Computer Science, Tsinghua University\\
{\tt\small yuh21@mails.tsinghua.edu.cn, xingxuanzhang@hotmail.com, xrz199721@gmail.com}\\ {\tt\small liujiashuo77@gmail.com, heyuethu@mail.tsinghua.edu.cn, cuip@tsinghua.edu.cn}
}

\begin{document}
\maketitle
\begin{abstract}
Domain generalization aims to solve the challenge of Out-of-Distribution (OOD) generalization by leveraging common knowledge learned from multiple training domains to generalize to unseen test domains. To accurately evaluate the OOD generalization ability, it is required that test data information is unavailable. However, the current domain generalization protocol may still have potential test data information leakage. 
This paper examines the risks of test data information leakage from two aspects of the current evaluation protocol: supervised pretraining on ImageNet and oracle model selection. We propose modifications to the current protocol that we should employ self-supervised pretraining or train from scratch instead of employing the current supervised pretraining, and we should use multiple test domains. These would result in a more precise evaluation of OOD generalization ability. We also rerun the algorithms with the modified protocol and introduce new leaderboards to encourage future research in domain generalization with a fairer comparison. 
\end{abstract}    
\section{Introduction}
\label{sec:intro}

The performance of traditional machine learning algorithms heavily depends on the assumption that the training and test data are independent and identically distributed (IID). However, in wild environments, the test distribution often differs significantly from the training distribution. This mismatch can lead to spurious correlations, ultimately causing the machine learning models to perform poorly and become unstable. Unfortunately, this limitation severely hinders the use of machine learning in high-risk domains like autonomous driving~\cite{huval2015empirical,levinson2011towards}, medical treatment~\cite{kukar2003transductive}, and law~\cite{berk2021fairness}. 

In recent years, there has been a growing interest among researchers in addressing the Out-of-Distribution (OOD) generalization problem, where the IID assumption does not stand~\cite{shen2021towards,yu2024survey}. This issue is addressed by various branches of research, such as invariant learning~\cite{arjovsky2019invariant, krueger2021out, creager2021environment, liu2021heterogeneous,liu2021integrated}, distributionally robust optimization~\cite{sinha2018certifying, duchi2021learning, volpi2018generalizing,liu2021stable,liu2022distributionally}, stable learning~\cite{kuang2020stable,shen2020stable,xu2021stable,yu2023stable,he2020learning,cui2022stable}, and domain generalization~\cite{zhou2022domain, wang2022generalizing, zhou2020domain, li2018domain, zhang2021deep}. 
Of these, domain generalization assumes the heterogeneity of training data. Specifically, domain generalization attempts to learn common or causal knowledge from multiple training domains to develop a model capable of generalizing to unseen test data. 

Despite a quantity of interesting and instructive works in domain generalization, previously they do not have a common standard training and evaluation protocol. Aware of this problem, DomainBed \cite{gulrajani2020search} proposes a framework for the hyperparameter search and model selection of domain generalization. It also sets a standard for experimental details like the model backbone, data split, data augmentation, etc. This unifies the protocol of domain generalization for subsequent works to follow. 
Nevertheless, conflicts remain between the current standard protocol and the accurate and reliable evaluation of OOD generalization ability. Since domain generalization is depicted as the ability to learn a model from diverse training domains that can generalize to \textbf{unseen/unknown} test data \cite{wang2022generalizing, zhou2022domain}, we should try to mitigate possible test data information leakage for a more precise evaluation of the OOD generalization ability.

In the current protocol, two key factors show the potential risk of test data information leakage. We make two recommendations for fairer and more accurate evaluation.

\begin{recommendation}
    Domain generalization algorithms should adopt self-supervised pretrained weights or random weights as initialization when evaluated and compared with each other.
\end{recommendation}
Most domain generalization algorithms take advantage of ImageNet supervised pretrained weights \cite{deng2009imagenet,gulrajani2020search} for better performance and faster convergence. Yet this introduces the information on both images and category labels in ImageNet, which may bear a resemblance to the test domain.  
Through comprehensive experiments, we demonstrate that more utilization of supervised pretrained weights and less utilization of training data can contribute to higher test domain performance under many common settings of domain generalization. This reveals that the ImageNet supervised pretrained weights may play a leading role in the test domain performance. Thus the accurate evaluation of OOD generalization is violated since the test domain performance does not really come from the generalization from training domains to test domains, for which most domain generalization algorithms are designed, but from the utilization of the supervised pretrained weights. 
We also demonstrate that when a test domain is quite similar to the ImageNet dataset, such a phenomenon becomes most evident while it does not occur if the test domain is rather different from ImageNet. This further confirms the test data information leakage. 

To address such an issue, it is safest to train from scratch to purely evaluate domain generalization. However, on one hand, with the remarkable development and broad application of pretrained models these days \cite{brown2020language,radford2021learning}, it is too limited and not common practice to train from scratch in real applications without benefiting from pretraining. On the other hand, most of commonly used domain generalization datasets like PACS \cite{li2017deeper} and VLCS \cite{khosla2012undoing} are not large enough to support training from scratch. 
Thus we investigate different pretrained methods and model backbones towards a set of pretrained weights with which there is less test data information leakage and we can still conduct a relatively accurate evaluation of OOD generalization. Based on our experimental findings, we suggest that self-supervised pretrained weights are a good alternative. 

\begin{recommendation}
    Domain generalization algorithms should be evaluated on multiple test domains. 
\end{recommendation}
For each trained model, domain generalization algorithms are typically evaluated on a single test domain. Before DomainBed, the choice of hyperparameters is not well specified, and there is a chance that model selection is conducted with the help of test data, i.e., the oracle model selection~\cite{gulrajani2020search}. This can introduce information leakage from the test data and undermine the validity of the evaluation. Even following the standard protocol of DomainBed, such a possibility still exists since the search space of hyperparameters could be pre-selected with the information of oracle data and fixed apriori for the DomainBed model selection pipeline. 
Moreover, the ultimate goal of domain generalization is to develop models that can generalize well to a wide range of unseen domains in real-world applications instead of tuning a set of hyperparameters for one single test domain. The current protocol allows models to select different hyperparameters for each test domain, which may not reflect the real-world scenario and could be inconsistent with the original purpose of domain generalization \cite{zhang2023nico++}. 
We suggest that we should evaluate algorithms on multiple test domains for each trained model since we empirically demonstrate that by doing so, the potential leakage from oracle model selection can be greatly mitigated. 

\paragraph{New leaderboards} Based on the aforementioned recommendations, we have conducted a re-evaluation of ten representative domain generalization algorithms following the revised protocol and presented three sets of new leaderboards. For ResNet50~\cite{he2016deep} that is employed in the current protocol, we provide leaderboards with MoCo-v2~\cite{chen2020improved} pretraining across all commonly used datasets, and leaderboards with no pretraining on large-scale datasets like DomainNet~\cite{peng2019moment} and NICO++~\cite{zhang2023nico++}. In addition, to support comparisons on more advanced network architectures like vision transformers, we also provide leaderboards for ViT-B/16~\cite{dosovitskiy2020image} with MoCo-v3~\cite{chen2021mocov3} pretraining. 
Combined with our previous analyses, the change in rankings of algorithms between the new leaderboard and the old one also implies that we are taking risks to improperly evaluate and rank existing methods with the current evaluation protocol. We believe the revised protocol and the leaderboards will stimulate future research in the field of domain generalization with more precise evaluation.


\section{Rethinking the Evaluation Protocol}

In this section, we will rethink the current evaluation protocol through comprehensive experimental analyses. 
First we review the definition of domain generalization. 

\begin{definition}[Domain Generalization]
\label{def:dg}
Given $M$ different training domains $\{S_1, S_2, ..., S_M\}$ where $S_j=\{x_i^{(j)},y_i^{(j)}\}_{i=1}^{n_j}$ is sampled from $P_j(X,Y)$. The goal is to learn a function $f$ that predicts well on the \textbf{unseen test domain} $S_{te}$. $S_{te}$ should be different from the $M$ training domains. With $l$ denoting the loss function, the optimization target is: 
\begin{equation}
    {\rm min}_f \mathbb{E}_{(x,y)\sim P_{te}}[l(f(x),y)]
\end{equation}
\label{eq:dg}
\end{definition}

Since the test domain is required to be unseen \cite{wang2022generalizing, zhou2022domain}, we should try to decrease the potential risk of leaking test data information to accurately evaluate the OOD generalization ability of algorithms. 

Although DomainBed has established a standard protocol for researchers of domain generalization to follow, there are still some defects hindering the accurate and fair evaluation of OOD generalization ability. Specifically, \cref{def:dg} does not explicitly address the use of pretrained weights, and the optimization target Equation \ref{eq:dg} is defined for a single test domain $P_{te}$. We will discuss these issues in detail in the following sections. 

We briefly introduce the domain generalization benchmark datasets we will use in our experiments.

\begin{itemize}
    \item PACS \cite{li2017deeper}: consists of 4 domains to depict the distribution shift as a change of style: photo, art\_painting, cartoon, sketch. It comprises 9,991 samples with 7 classes. 
    \item VLCS \cite{khosla2012undoing}: is collected from 4 different datasets corresponding to four domains:  Pascal VOC 2007, LabelMe, Caltech, SUN09. It contains 10,729 real photo examples with 5 common classes. It depicts the distribution shift with dataset bias. 
    \item OfficeHome \cite{venkateswara2017deep}: comprises 4 domains: art, clipart, product, real, with 65 classes and 15,588 examples. It also depicts the style transfer. 
    \item Terra Incognita \cite{beery2018recognition}: comprises 4 domains: L38, L43, L46, L100, with 10 classes and 24,788 examples. Its distribution shift is characterized by different locations when taking photos. 
    \item DomainNet \cite{peng2019moment}: comprises 6 domains: clipart, infograph, painting, quickdraw, real, sketch, also characterized by style shift. It is a relatively large dataset with 586,575 samples and 345 classes in total. 
    \item NICO++ \cite{zhang2023nico++}: comprises 6 publicly available domains: autumn, rock, dim, grass, outdoor, water. This part contains 88,866 examples with 60 classes. It is also a relatively large dataset that controls the distribution shift through the change of background contexts. 
\end{itemize}

\begin{table}[tbp]
\caption{Results of linear-probing (LP) and fine-tuning (FT) with supervised pretrained ResNet-50 on commonly used domain generalization datasets. }
\label{table:supervised}
\resizebox{\linewidth}{!}{%
\begin{tabular}{@{}c|cccc|c@{}}
\toprule
PACS       & P                 & A        & C                 & S                 & Avg               \\ \midrule
LP         & \textbf{97.7±0.1} & 71.8±1.6 & 53.8±1.8          & 45.9±1.7          & 67.3±0.3          \\
FT         & 97.4±0.1          & 86.1±0.9 & 80.4±1.4          & 77.1±2.5          & 85.3±0.6          \\ \midrule
VLCS       & V                 & L        & C                 & S                 & Avg               \\ \midrule
LP         & \textbf{77.2±1.6} & 58.1±0.6 & \textbf{97.4±0.4} & 71.4±1.1          & 76.0±0.6          \\
FT         & 73.5±1.5          & 66.3±0.9 & 96.9±1.1          & 71.7±1.5          & 77.1±1.0          \\ \midrule
OfficeHome & A                 & C        & P                 & R                 & Avg               \\ \midrule
LP         & \textbf{64.0±0.4} & 50.3±0.3 & \textbf{77.7±0.5} & \textbf{79.7±0.2} & \textbf{67.9±0.2} \\
FT         & 61.1±0.6          & 51.1±0.3 & 73.9±0.5          & 75.7±0.7          & 65.5±0.2          \\ \midrule
TerraInc   & L38               & L43      & L46               & L100              & Avg               \\ \midrule
LP         & \textbf{43.4±5.4} & 36.6±0.9 & 32.4±0.9          & 37.9±0.2          & 37.6±1.7          \\
FT         & 43.2±2.4          & 56.1±0.2 & 38.4±5.7          & 54.8±5.9          & 48.1±3.1          \\ \bottomrule
\end{tabular}%
}

\end{table}

\subsection{Pretraining}
\label{sec:pretrain}

When the most widely used domain generalization benchmark dataset PACS is released \cite{li2017deeper}, the authors proposed a baseline "Deep-All", an AlexNet \cite{krizhevsky2017imagenet} supervised pretrained on ImageNet and optimized using ERM \cite{vapnik1991principles}. 
To ensure comparability with prior research, subsequent studies have consistently followed this practice, even as the backbone architecture has evolved from AlexNet \cite{li2017deeper} to ResNet-18 \cite{carlucci2019domain, matsuura2020domain, nam2021reducing, huang2020self} and more recently to ResNet-50 \cite{gulrajani2020search, cha2021swad, rame2022fishr}. 
It is worth noting that bringing in extra knowledge beyond training data may have a potential negative effect of test data information leakage, hurting the accurate evaluation of OOD generalization. For instance, VLCS \cite{khosla2012undoing}, TerraInc \cite{beery2018recognition} and NICO++ \cite{zhang2023nico++} comprise entirely real photos, which resemble the pretrained dataset ImageNet. Similarly, some domains in PACS \cite{li2017deeper}, OfficeHome \cite{venkateswara2017deep}, and DomainNet \cite{peng2019moment} also include real photos that share similar characteristics with ImageNet. When using these data as test domains, there could be potential leakage of test data information from the pretrained weights.

\begin{figure}[t]
  \centering
  \includegraphics[width=0.9\linewidth]{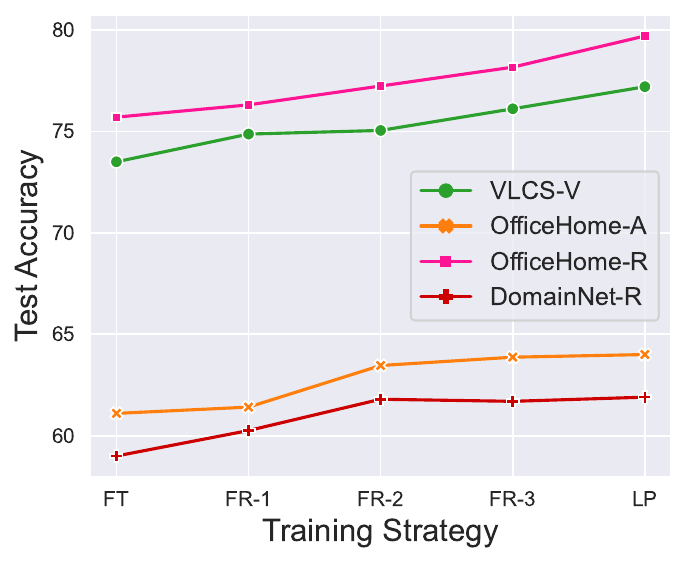}
  \caption{Test domain accuracy with the growing number of frozen layers when using ImageNet pretrained weights. Test domains include PASCAL from VLCS, Art and Real from OfficeHome, and Real from DomainNet. }
  \label{fig:freeze}
\end{figure}

\begin{table}[h]
\centering
\caption{Relationship between dissimilarity of pretrained data and test domains, and the phenomenon of LP outperforming FT. The "Gap" is calculated as $\frac{{\rm Acc}_{\rm LP}-{\rm Acc}_{\rm FT}}{{\rm Acc}_{\rm FT}}$. }
\label{table:fid}
\resizebox{0.75\linewidth}{!}{%
\begin{tabular}{@{}c|cccc@{}}
\toprule
PACS            & P      & A      & C      & S      \\ \midrule
 Gap & 0.004  & -0.167 & -0.331 & -0.405 \\
FID             & 103.75 & 128.07 & 148.52 & 209.68 \\ \midrule
OfficeHome      & A      & C      & P      & R      \\ \midrule
 Gap & 0.047  & -0.017 & 0.051  & 0.054  \\
FID             & 62.22  & 92.33  & 78.26  & 56.96  \\ \bottomrule
\end{tabular}%
}
\end{table}

To confirm our concerns, we conduct experiments by comparing the effect of pretrained weights and training data on the test domain performance. 
We basically follow the protocol of DomainBed \cite{gulrajani2020search} with an ImageNet supervised pretrained ResNet-50. We compare two paradigms: LP and FT. LP stands for linear-probing where only the last layer is updated. 
In real applications where the cost of fine-tuning the whole network is too high to be afforded, LP is used to save computational resources and guarantee the generalization ability to some extent. 
FT stands for fine-tuning the whole network. In general, LP relies more on the pretrained features and FT relies more on the fine-tuned training data. For both LP and FT, we employ the standard ERM to fine-tune models on the data of training domains. 
In traditional computer vision tasks, if the training data is sufficient for the network to be trained, FT is expected to yield superior results as compared to LP, as it is better equipped to leverage the training data \cite{chen2020big, chen2021empirical}. However, as evident from \cref{table:supervised}, LP performs comparably with FT under many settings, in some of which simple LP even just outperforms FT. For VLCS and OfficeHome, LP outperforms FT or performs comparably with LP across almost all settings. For PACS, when using real photos as the test domain, LP outperforms FT. For TerraInc, LP also outperforms FT in the domain L38. 
Such a phenomenon can also be observed in DomainNet and NICO++, which is left in \cref{supp:supervised}. 
As LP only updates the last layer, it strongly relies on the pretrained weights to generalize on the test domains, while FT relies more on the training domains and less on the pretrained weights than LP. 
The above results imply that the test domain performance under many settings is mostly attributed to the information from the pretrained weights rather than the information from the training data.

In order to further investigate the phenomenon that LP outperforms FT, we vary the number of frozen layers in ResNet-50, which we consider as a 4-layer structure. Specifically, we fine-tune the network by freezing the first 1, 2, and 3 layers, respectively. 
\cref{fig:freeze} shows that as the number of frozen layers increases, indicating higher utilization of pretrained weights and lower utilization of training domain data, the test domain accuracies also increase. 
Besides, we calculate Fréchet Inception Distance (FID) \cite{heusel2017gans} between some domains and ImageNet in \cref{table:fid} and find that in most cases, a smaller FID is accompanied by a larger value of $\frac{{\rm Acc}_{\rm LP}-{\rm Acc}_{\rm FT}}{{\rm Acc}_{\rm FT}}$, where ${\rm Acc}_{\rm LP}$ and ${\rm Acc}_{\rm FT}$ represent the corresponding test accuracy. This implies that the phenomenon of LP outperforming FT can really serve as evidence of test data information leakage, i.e. a higher similarity of pretrained data to test data. 

These findings raise concerns about the fairness of comparing different algorithms with the existence of test data information leakage from supervised pretrained weights. 
If an algorithm can improve the utilization of the pretrained weights instead of the true OOD generalization ability from training domains to test domains (e.g. in some settings, simple LP already brings higher improvement than many domain generalization algorithms), it may generate comparable or even better test domain performance than the algorithms that are better at true OOD generalization. 
To solve this issue, the safest choice is to train from scratch when evaluating domain generalization algorithms. Nevertheless, as pretrained models have achieved rapid growth in recent years, pretraining is commonly used to improve performance without incurring additional time or financial costs. It is somewhat limited to evaluate algorithms by training from scratch which may lead to a large gap between evaluation and real-world model deployment. 

Therefore, we explore alternatives of pretraining that mitigate the risk of leakage, from perspectives of backbone architectures and pretraining methods. For changing backbone architectures, we conduct experiments with supervised pretrained ResNet-18 and ViT-B/16~\cite{dosovitskiy2020image}. For changing pretraining methods, we try several self-supervised pretraining including MoCo~\cite{he2020momentum}, MoCo-v2~\cite{chen2020improved}, SimCLR~\cite{chen2020simple}, SimCLR-v2~\cite{chen2020big} for ResNet-50, and MoCo-v3~\cite{chen2021mocov3} for ViT-B/16. 
From results in \cref{table:officehome} that take OfficeHome as an example, we find that after changing from ResNet-50 to ResNet-18 (decreasing model capacity) or ViT-B/16 (increasing model capacity), the phenomenon of LP outperforming FT still exists. However, by changing from supervised pretraining to self-supervised pretraining, the phenomenon of LP outperforming FT disappears, for both ResNet-50 and ViT-B/16. Similar observations are made for other datasets, which we leave in \cref{supp:various}. These results demonstrate that changing model architectures does not help, but changing to self-supervised pretraining greatly helps in mitigating potential risks of test data information leakage. Since self-supervised pretraining only utilizes images compared with supervised pretraining utilizing both images and category labels, these results adhere to our intuition that self-supervised pretraining may bear less leakage. 

Overall, we suggest that we should use self-supervised pretrained weights or train from scratch to mitigate the potential risk of test information leakage for fairer evaluation.

\begin{table}[t]
\centering
\caption{Results of linear-probing (LP) and fine-tuning (FT) with different pretraining methods of different backbone architectures on OfficeHome. }
\label{table:officehome}
\resizebox{\linewidth}{!}{%
\begin{tabular}{@{}ccc|cccc|c@{}}
\toprule
\multicolumn{3}{c|}{OfficeHome}                                                                          & A                 & C        & P                 & R                 & Avg               \\ \midrule
\multicolumn{1}{c|}{\multirow{2}{*}{ResNet-18}}  & \multicolumn{1}{c|}{\multirow{2}{*}{Supervised}} & LP & \textbf{57.4±0.3} & 46.0±0.2 & \textbf{71.3±0.2} & \textbf{74.5±0.3} & \textbf{62.3±0.1} \\
\multicolumn{1}{c|}{}                            & \multicolumn{1}{c|}{}                            & FT & 50.8±0.5          & 46.5±1.1 & 67.4±0.4          & 69.2±0.4          & 58.5±0.3          \\ \midrule
\multicolumn{1}{c|}{\multirow{10}{*}{ResNet-50}} & \multicolumn{1}{c|}{\multirow{2}{*}{Supervised}} & LP & \textbf{64.0±0.4} & 50.3±0.3 & \textbf{77.7±0.5} & \textbf{79.7±0.2} & \textbf{67.9±0.2} \\
\multicolumn{1}{c|}{}                            & \multicolumn{1}{c|}{}                            & FT & 61.1±0.6          & 51.1±0.3 & 73.9±0.5          & 75.7±0.7          & 65.5±0.2          \\ \cmidrule(l){2-8} 
\multicolumn{1}{c|}{}                            & \multicolumn{1}{c|}{\multirow{2}{*}{MoCo}}       & LP & 27.5±0.6          & 17.3±0.1 & 32.8±0.3          & 40.1±0.6          & 29.4±0.2          \\
\multicolumn{1}{c|}{}                            & \multicolumn{1}{c|}{}                            & FT & 45.3±0.7          & 36.9±1.1 & 62.4±0.8          & 64.2±0.1          & 52.2±0.6          \\ \cmidrule(l){2-8} 
\multicolumn{1}{c|}{}                            & \multicolumn{1}{c|}{\multirow{2}{*}{MoCo-v2}}    & LP & 41.5±0.4          & 25.3±0.2 & 49.9±0.4          & 56.7±0.2          & 43.4±0.1          \\
\multicolumn{1}{c|}{}                            & \multicolumn{1}{c|}{}                            & FT & 49.6±2.7          & 45.2±2.0 & 65.8±1.3          & 68.6±0.2          & 57.3±0.3          \\ \cmidrule(l){2-8} 
\multicolumn{1}{c|}{}                            & \multicolumn{1}{c|}{\multirow{2}{*}{SimCLR}}     & LP & 9.7±0.2           & 7.9±0.0  & 12.6±0.3          & 17.0±0.3          & 11.8±0.1          \\
\multicolumn{1}{c|}{}                            & \multicolumn{1}{c|}{}                            & FT & 24.6±0.6          & 26.3±1.5 & 44.3±1.0          & 41.7±0.6          & 34.2±0.4          \\ \cmidrule(l){2-8} 
\multicolumn{1}{c|}{}                            & \multicolumn{1}{c|}{\multirow{2}{*}{SimCLR-v2}}  & LP & 6.3±0.3           & 5.8±0.2  & 7.5±0.2           & 10.2±0.2          & 7.5±0.1           \\
\multicolumn{1}{c|}{}                            & \multicolumn{1}{c|}{}                            & FT & 42.9±2.7          & 43.9±1.8 & 63.3±0.7          & 64.9±0.4          & 53.8±0.3          \\ \midrule
\multicolumn{1}{c|}{\multirow{4}{*}{ViT-B/16}}   & \multicolumn{1}{c|}{\multirow{2}{*}{Supervised}} & LP & \textbf{72.7±0.4} & 57.0±0.2 & \textbf{82.7±0.1} & \textbf{83.8±0.1} & \textbf{74.1±0.1} \\
\multicolumn{1}{c|}{}                            & \multicolumn{1}{c|}{}                            & FT & 71.1±1.2          & 59.1±0.6 & 80.6±0.9          & 83.3±0.4          & 73.5±0.5          \\ \cmidrule(l){2-8} 
\multicolumn{1}{c|}{}                            & \multicolumn{1}{c|}{\multirow{2}{*}{MoCo-v3}}    & LP & 63.1±0.3          & 38.8±0.2 & 69.1±0.3          & 73.0±0.3          & 61.0±0.2          \\
\multicolumn{1}{c|}{}                            & \multicolumn{1}{c|}{}                            & FT & 64.7±0.6          & 54.1±0.5 & 74.8±0.7          & 78.3±0.7          & 68.0±0.4          \\ \bottomrule
\end{tabular}%
}
\end{table}

\begin{table}[]
  \caption{Results of increasing the number of test domains for mitigating test information leakage from oracle model selection for DomainNet and NICO++. }
  \label{oracle}

\vspace{-3pt}

\resizebox{\linewidth}{!}{%
\begin{tabular}{@{}cc|ccc|c|cc@{}}
\toprule
\multicolumn{2}{c|}{DomainNet}                     & quickdraw & real  & sketch & Avg   & \multicolumn{2}{c}{leakage} \\ \midrule
\multicolumn{2}{c|}{IID}                           & 11.09     & 60.75 & 47.98  & 39.94 & 0.00         & /            \\ \midrule
\multicolumn{1}{c|}{\multirow{3}{*}{Oracle}} & K=1 & 11.42     & 62.57 & 48.30  & 40.76 & 0.82         & -0\%         \\
\multicolumn{1}{c|}{}                        & K=2 & 10.95     & 62.54 & 48.03  & 40.51 & 0.57         & -31\%        \\
\multicolumn{1}{c|}{}                        & K=3 & 10.81     & 62.54 & 48.07  & 40.47 & 0.53         & -35\%        \\ \bottomrule
\end{tabular}%
}

\vspace{3pt}

\resizebox{\linewidth}{!}{%
\begin{tabular}{@{}cc|ccc|c|cc@{}}
\toprule
\multicolumn{2}{c|}{DomainNet}                     & clipart & infograph & painting & Avg   & \multicolumn{2}{c}{leakage} \\ \midrule
\multicolumn{2}{c|}{IID}                           & 58.23   & 19.20     & 47.49    & 41.64 & 0.00         & /            \\ \midrule
\multicolumn{1}{c|}{\multirow{3}{*}{Oracle}} & K=1 & 58.23   & 19.23     & 47.67    & 41.71 & 0.07         & -0\%         \\
\multicolumn{1}{c|}{}                        & K=2 & 58.19   & 19.22     & 47.67    & 41.69 & 0.05         & -24\%        \\
\multicolumn{1}{c|}{}                        & K=3 & 58.15   & 19.23     & 47.67    & 41.68 & 0.04         & -38\%        \\ \bottomrule
\end{tabular}%
}

\vspace{3pt}

\resizebox{\linewidth}{!}{%
\begin{tabular}{@{}cc|ccc|c|cc@{}}
\toprule
\multicolumn{2}{c|}{NICO++}                        & grass & outdoor & water & Avg   & \multicolumn{2}{c}{leakage} \\ \midrule
\multicolumn{2}{c|}{IID}                           & 80.10 & 74.69   & 68.06 & 74.28 & 0.00         & /            \\ \midrule
\multicolumn{1}{c|}{\multirow{3}{*}{Oracle}} & K=1 & 81.19 & 75.58   & 69.85 & 75.54 & 1.26         & -0\%         \\
\multicolumn{1}{c|}{}                        & K=2 & 80.54 & 74.98   & 69.85 & 75.12 & 0.84         & -33\%        \\
\multicolumn{1}{c|}{}                        & K=3 & 80.47 & 75.58   & 68.97 & 75.01 & 0.72         & -42\%        \\ \bottomrule
\end{tabular}%
}

\vspace{3pt}

\resizebox{\linewidth}{!}{%
\begin{tabular}{@{}cc|ccc|c|cc@{}}
\toprule
\multicolumn{2}{c|}{NICO++}                        & autumn & rock  & dim   & Avg   & \multicolumn{2}{c}{leakage} \\ \midrule
\multicolumn{2}{c|}{IID}                           & 78.94  & 79.26 & 71.94 & 76.71 & 0.00         & /            \\ \midrule
\multicolumn{1}{c|}{\multirow{3}{*}{Oracle}} & K=1 & 79.44  & 79.58 & 72.58 & 77.20 & 0.49         & -0\%         \\
\multicolumn{1}{c|}{}                        & K=2 & 78.92  & 79.33 & 72.58 & 76.94 & 0.23         & -53\%        \\
\multicolumn{1}{c|}{}                        & K=3 & 79.44  & 79.58 & 71.42 & 76.81 & 0.10         & -79\%        \\ \bottomrule
\end{tabular}%
}
\vspace{-12pt}
\end{table}

\subsection{Oracle Model Selection}

Model selection has emerged as a crucial problem for OOD generalization. In the context of IID generalization, training data, validation data, and test data are typically drawn from the same distribution, so the generalization performances on validation data and test data are generally consistent. However, in the case of domain generalization, test distribution differs from training distribution. Since test data should be unknown, naturally the validation data should also come from the same distribution as training data does, thus the consistency between validation data performance and test data performance cannot be guaranteed. When evaluating on public benchmarks, despite not being reasonable, it is possible to exploit test data for hyperparameter tuning and model selection, referred to as "oracle" model selection. This serves as another form of test data information leakage. 
DomainBed \cite{gulrajani2020search} has proved that such an oracle model selection method outperforms the standard model selection strategy that utilizes a validation set sampled from the same distribution as training data. While DomainBed has integrated the validation process into the framework to reduce the possibility of using test data, there are still some degrees of freedom. For example, the hyperparameter search space can be customized and narrowed in order to reduce the computational cost of the evaluation process (In the current standard protocol of DomainBed, each setting randomly generates 20 hyperparameter sets for 3 random seeds each, leading to the requirement of training 60 models). 

To further address this issue, we observe that the current protocol of domain generalization, as defined by Equation \ref{eq:dg}, only considers a single test domain for each trained model. 
Though the test domain varies across different settings, it is fixed for a single setting.
For example, when evaluating on DomainNet with domains \emph{real} and \emph{clipart}, we do not directly train a single model for the two settings with each domain treated as the test domain. 
Instead, we train two separate models using different training data comprising domains other than the test domain. The two hyperparameter search and model selection processes are also performed independently for each model. As there is only one test domain, the room for increasing the test performance through oracle model selection is relatively large. Intuitively, using oracle model selection to "fit" the distribution of multiple test domains is usually harder than "fit" the distribution of a single test domain. Thus we consider introducing multiple test domains to alleviate the test data information leakage from the oracle model selection method.

We conduct experiments to confirm the above intuition. We adopt DomainNet and NICO++ due to their relatively larger number of domains. We split the domains of each dataset into two groups. For DomainNet, we split them into (quickdraw, real, sketch) and (clipart, infograph, painting). For NICO++, we split them into (grass, outdoor, water) and (autumn, rock, dim). For each dataset, we train on each group respectively and test on the other group. Validation data is randomly split from training data. For IID model selection, we choose the test accuracy corresponding to the hyperparameters with the highest accuracy on validation data. For oracle model selection, we directly choose the highest test accuracy across all hyperparameter sets. We vary the number of test domains $K$ and compute the test accuracy of each test domain through the average of its accuracy across every combination of $K$ test domains that includes this test domain. For example, for the domain \emph{autumn} belonging to the group (autumn, rock, dim), for $K=2$, we calculate the average of domain autumn's accuracy when using (autumn, rock) and (autumn, dim) for oracle model selection respectively. 
We quantify the difference in test accuracy between IID model selection and oracle model selection as the possible "leakage". 
\cref{oracle} shows that increasing the number of test domains does help mitigate the possible leakage from oracle model selection. For DomainNet, when $K=2$, the leakage can already be reduced by about 30 percent. For NICO++, the leakage can be reduced by about half when increasing $K$. If there are datasets that can support the evaluation of more test domains, such leakage can be further reduced. 
Overall, we recommend using multiple test domains to alleviate the possible information leakage from oracle model selection.

\section{New Leaderboards}
The above analyses suggest that to reduce the risk of test data information leakage and ensure accurate evaluation of OOD generalization, it is advisable to use self-supervised pretrained weights or train from scratch, and to increase the number of test domains. We have accordingly introduced these two modifications to the DomainBed protocol and present new leaderboards that are compared with the old one which follows the current DomainBed protocol. 

\subsection{Experimental settings}

\paragraph{Protocol modifications} 

For self-supervised pretraining investigated in \cref{table:officehome}, we choose MoCo-v2~\cite{chen2020improved} pretrained ResNet-50 for the new evaluation protocol since it outperforms other self-supervised pretrained weights when using ResNet-50 as backbones. To support comparisons on more advanced architectures like ViT, we employ MoCo-v3~\cite{chen2021mocov3} pretrained ViT-B/16. 
Since there are only 4 domains in datasets other than DomainNet and NICO++, we still employ leave-one-domain-out strategy. For DomainNet and NICO++ with 6 domains each, we divide them into 3 groups: (clipart, infograph), (painting, quickdraw), (real, sketch) for DomainNet, and (autumn, rock), (dim, grass), (outdoor, water) for NICO++. We employ the leave-one-group-out strategy so that each time we have 2 domains for testing. 
Due to observations that there are relatively large random fluctuations in the results of TerraInc both in our experiments and in DomainBed, we do not adopt it in our new leaderboards. 
Besides, we present leaderboards without pretraining for DomainNet and NICO++ in \cref{supp:none}, since only they are large enough for ResNet and ViT to be sufficiently trained on from scratch. 

\paragraph{Algorithms}

We test 10 algorithms following the modified protocol: ERM \cite{vapnik1991principles}, SWAD \cite{izmailov2018averaging, cha2021swad}, RSC \cite{huang2020self}, GroupDRO \cite{sagawa2019distributionally}, Fishr \cite{rame2022fishr}, CORAL \cite{sun2016return, sun2016deep}, MMD~\cite{li2018domain}, SagNet \cite{nam2021reducing}, IRM \cite{arjovsky2019invariant}, Mixup \cite{zhang2018mixup, yan2020improve}.

ERM directly optimizes sample averaged loss, typically used in traditional machine learning tasks with IID assumption. 
Among these algorithms, Fishr, CORAL, MMD, SagNet, and IRM aim to achieve some form of invariance across domains. Mixup aims to enhance the diversity of the training data. 
Some of the algorithms were originally developed for other areas, such as CORAL and inter-domain Mixup for domain adaptation, IRM for invariant learning, GroupDRO for subpopulation shift, and SWAD adapted from the optimizer seeking flat minima. 

\paragraph{Other details}

For PACS, VLCS, and OfficeHome, we set the number of iterations as 5,000 following DomainBed~\cite{gulrajani2020search}. For DomainNet, we set it as 15,000 following \citet{cha2021swad} since the training loss has not converged yet at the iteration of 5,000. For NICO++, we set it as 10,000. 
To reduce the computational cost of the current DomainBed protocol, for each setting, namely each combination of an algorithm and a pair of training and test domains, we randomly search the hyperparameters over a predefined distribution with 10 trials (instead of 20 trials in DomainBed). We use the selected best hyperparameters to run 2 more times with different random seeds (instead of conducting 3 independent hyperparameter searches for each random seed). The total cost is training 12 models (instead of 60 models). 
As for the predefined hyperparameter search space, we change the search space of MMD gamma from log uniform distribution of $10^{{\rm uniform}(-1,1)}$ to that of $10^{{\rm uniform}(-2,0)}$ otherwise we observe that training loss will not decrease. 
Besides, in all our experiments, we use the Adam optimizer \cite{kingma2014adam} with a Cosine Annealing Scheduler. 
For details of training from scratch, we put them in \cref{supp:none}. 
For other details, such as data augmentation and data split, we directly follow DomainBed. 

\begin{table*}[]
  \caption{Leaderboards comparison between the current DomainBed protocol and the modified evaluation protocol of adopting self-supervised pretraining and using multiple test domains. }
  \label{leaderboard}
  \centering
\resizebox{0.94\textwidth}{!}{%
\begin{tabular}{@{}ccccccccc@{}}
\toprule
\multicolumn{9}{c}{Old   leaderboard: Supervised pretrained ResNet-50}                                                                         \\ \midrule
\multicolumn{1}{c|}{Algorithm} & PACS     & VLCS     & OfficeHome & NICO++   & \multicolumn{1}{c|}{DomainNet} & Average & Ranking & $\Delta R$ \\ \midrule
\multicolumn{1}{c|}{ERM}       & 85.5±0.2 & 77.5±0.4 & 66.5±0.3   & 77.5±0.1 & \multicolumn{1}{c|}{40.9±0.1}  & 69.6    & 6       & -          \\
\multicolumn{1}{c|}{SWAD}      & 88.1±0.1 & 79.1±0.1 & 70.6±0.2   & 80.2±0.1 & \multicolumn{1}{c|}{46.5±0.1}  & 72.9    & 1       & -          \\
\multicolumn{1}{c|}{RSC}       & 85.2±0.9 & 77.1±0.5 & 65.5±0.9   & 78.1±0.2 & \multicolumn{1}{c|}{38.9±0.5}  & 69.0    & 7       & -          \\
\multicolumn{1}{c|}{GroupDRO}  & 84.4±0.8 & 76.7±0.6 & 66.0±0.7   & 77.6±0.4 & \multicolumn{1}{c|}{33.3±0.2}  & 67.6    & 8       & -          \\
\multicolumn{1}{c|}{Fishr}     & 85.5±0.4 & 77.8±0.1 & 67.8±0.1   & 78.4±0.1 & \multicolumn{1}{c|}{41.7±0.0}  & 70.2    & 3       & -          \\
\multicolumn{1}{c|}{CORAL}     & 86.2±0.3 & 78.8±0.6 & 68.7±0.3   & 79.3±0.1 & \multicolumn{1}{c|}{41.5±0.1}  & 70.9    & 2       & -          \\
\multicolumn{1}{c|}{MMD}       & 84.6±0.5 & 77.5±0.9 & 66.3±0.1   & 77.3±0.2 & \multicolumn{1}{c|}{23.4±9.5}  & 65.8    & 10      & -          \\
\multicolumn{1}{c|}{SagNet}    & 86.3±0.2 & 77.8±0.5 & 68.1±0.1   & 78.6±0.2 & \multicolumn{1}{c|}{40.3±0.1}  & 70.2    & 4       & -          \\
\multicolumn{1}{c|}{IRM}       & 83.5±0.8 & 78.5±0.5 & 64.3±2.2   & 77.5±0.3 & \multicolumn{1}{c|}{33.9±2.8}  & 67.5    & 9       & -          \\
\multicolumn{1}{c|}{Mixup}     & 84.6±0.6 & 77.4±0.6 & 68.1±0.3   & 78.9±0.0 & \multicolumn{1}{c|}{39.2±0.1}  & 69.6    & 5       & -          \\ \bottomrule
\end{tabular}%
}

\vspace{5pt}

\resizebox{0.94\textwidth}{!}{%
\begin{tabular}{@{}ccccccccc@{}}
\toprule
\multicolumn{9}{c}{New   leaderboard: MoCo-v2 pretrained ResNet-50}                                                                             \\ \midrule
\multicolumn{1}{c|}{Algorithm} & PACS     & VLCS     & OfficeHome & NICO++   & \multicolumn{1}{c|}{DomainNet} & Average & Ranking & $\Delta {\rm R}$  \\ \midrule
\multicolumn{1}{c|}{ERM}       & 84.1±0.3 & 76.9±0.8 & 57.3±0.3   & 71.1±0.1 & \multicolumn{1}{c|}{39.3±0.3}  & 65.7    & 6       & -           \\
\multicolumn{1}{c|}{SWAD}      & 86.2±0.6 & 77.7±0.7 & 62.6±0.1   & 75.6±0.1 & \multicolumn{1}{c|}{42.9±0.1}  & 69.0    & 1       & -           \\
\multicolumn{1}{c|}{RSC}       & 85.9±1.9 & 78.5±0.3 & 60.3±0.3   & 74.2±0.5 & \multicolumn{1}{c|}{40.5±0.4}  & 67.9    & 2       & \textbf{+5} \\
\multicolumn{1}{c|}{GroupDRO}  & 83.0±0.5 & 76.5±0.8 & 57.4±0.8   & 71.2±0.2 & \multicolumn{1}{c|}{36.3±0.4}  & 64.9    & 10      & -2           \\
\multicolumn{1}{c|}{Fishr}     & 82.8±1.0 & 75.4±1.6 & 58.7±0.2   & 71.0±0.1 & \multicolumn{1}{c|}{39.3±0.3}  & 65.4    & 8       & \textbf{-5} \\
\multicolumn{1}{c|}{CORAL}     & 84.0±1.4 & 77.5±0.4 & 62.6±0.3   & 73.7±0.2 & \multicolumn{1}{c|}{41.2±0.1}  & 67.8    & 3       & -1           \\
\multicolumn{1}{c|}{MMD}       & 84.3±1.2 & 76.7±1.3 & 57.6±0.8   & 71.6±0.3 & \multicolumn{1}{c|}{39.3±0.3}  & 65.9    & 5       & \textbf{+5}          \\
\multicolumn{1}{c|}{SagNet}    & 82.5±1.1 & 75.5±0.8 & 59.2±1.0   & 70.2±0.3 & \multicolumn{1}{c|}{38.6±0.2}  & 65.2    & 9       & \textbf{-5} \\
\multicolumn{1}{c|}{IRM}       & 83.5±0.8 & 75.3±0.5 & 57.9±0.3   & 71.5±0.3 & \multicolumn{1}{c|}{39.1±0.1}  & 65.5    & 7       & +2          \\
\multicolumn{1}{c|}{Mixup}     & 82.5±1.6 & 76.1±0.7 & 60.5±0.2   & 73.3±0.2 & \multicolumn{1}{c|}{39.2±0.1}  & 66.3    & 4       & +1          \\ \bottomrule
\end{tabular}
}

\vspace{5pt}

\resizebox{0.94\textwidth}{!}{%
\begin{tabular}{@{}ccccccccc@{}}
\toprule
\multicolumn{9}{c}{New   leaderboard: MoCo-v3 pretrained ViT-B/16}                                                                              \\ \midrule
\multicolumn{1}{c|}{Algorithm} & PACS     & VLCS     & OfficeHome & NICO++   & \multicolumn{1}{c|}{DomainNet} & Average & Ranking & $\Delta {\rm R}$  \\ \midrule
\multicolumn{1}{c|}{ERM}       & 85.8±0.3 & 78.4±0.6 & 68.0±0.4   & 79.6±0.2 & \multicolumn{1}{c|}{47.4±0.2}  & 71.8    & 5       & +1          \\
\multicolumn{1}{c|}{SWAD}      & 88.1±0.3 & 79.0±0.1 & 71.4±0.4   & 80.8±0.0 & \multicolumn{1}{c|}{49.6±0.1}  & 73.8    & 1       & -           \\
\multicolumn{1}{c|}{RSC}       & 86.8±0.5 & 78.2±1.1 & 67.9±0.1   & 79.7±0.1 & \multicolumn{1}{c|}{47.3±0.1}  & 72.0    & 2       & \textbf{+5} \\
\multicolumn{1}{c|}{GroupDRO}  & 85.6±0.8 & 78.2±0.9 & 68.0±0.2   & 79.7±0.2 & \multicolumn{1}{c|}{44.9±0.1}  & 71.3    & 8       & -          \\
\multicolumn{1}{c|}{Fishr}     & 86.6±1.3 & 78.0±0.4 & 67.7±0.3   & 79.6±0.2 & \multicolumn{1}{c|}{47.2±0.1}  & 71.8    & 6       & -3 \\
\multicolumn{1}{c|}{CORAL}     & 86.7±0.9 & 78.1±0.6 & 67.8±0.5   & 79.5±0.1 & \multicolumn{1}{c|}{47.5±0.1}  & 71.9    & 3       & -1           \\
\multicolumn{1}{c|}{MMD}       & 86.4±0.9 & 64.4±4.3 & 67.2±0.2   & 69.7±1.0 & \multicolumn{1}{c|}{47.3±0.1}  & 67.0    & 10      & -          \\
\multicolumn{1}{c|}{SagNet}    & 85.6±1.0 & 78.0±0.5 & 66.9±0.2   & 79.2±0.2 & \multicolumn{1}{c|}{46.5±0.1}  & 71.2    & 9       & \textbf{-5} \\
\multicolumn{1}{c|}{IRM}       & 84.7±0.5 & 78.1±0.3 & 68.2±0.5   & 79.7±0.1 & \multicolumn{1}{c|}{47.3±0.1}  & 71.6    & 7       & +2          \\
\multicolumn{1}{c|}{Mixup}     & 83.4±0.8 & 78.2±0.8 & 70.0±0.1   & 79.8±0.2 & \multicolumn{1}{c|}{48.0±0.1}  & 71.9    & 4       & +1          \\ \bottomrule
\end{tabular}
}
\end{table*}

\subsection{Results}

\cref{leaderboard} presents the old leaderboard based on supervised pretrained ResNet-50 and the two new leaderboards. In the old leaderboard, the results of SWAD are from \citet{cha2021swad}, the results of Fishr are from \citet{rame2022fishr}, and the others are from the standard leaderboard maintained in the official code repository of DomainBed \cite{gulrajani2020search}. The last column $\Delta {\rm R}$ represents the change of ranking for the algorithm, where we mark the largest changes with bold type. 
The detailed leaderboards for each dataset are in \cref{supp:leaderboard}.

\paragraph{Performance rankings of some algorithms show great variations after applying the modified protocol. } 

For instance, RSC ranks 2nd in both new leaderboards with self-supervised pretraining, implying its effectiveness, but it fails to outperform ERM in the old leaderboard with supervised pretraining. 
For a recent SOTA algorithm Fishr proposed based on the current DomainBed protocol, it achieves a high test accuracy in the old leaderboard but fails to outperform ERM in the new ones. A similar conclusion stands for SagNet too. Coupled with our analyses in \cref{sec:pretrain}, this raises concerns about the preciseness and fairness of the current evaluation protocol. 

\paragraph{Rankings in self-supervised pretraining leaderboards are more consistent with leaderboards of training from scratch than supervised pretraining leaderboards. }

We calculate spearman rank correlation for algorithm performance in different leaderboards. The rank correlation between supervised pretraining and training from scratch is 0.261 while MoCo-v2 and MoCo-v3's rank correlations with training from scratch are 0.576 and 0.600 respectively. Besides, rank correlation between MoCo-v2 and MoCo-v3 is 0.794. This implies that evaluation based on self-supervised pretraining is more effective than the currently used supervised pretraining since it serves as a better surrogate of training from scratch. This also confirms that self-supervised pretraining helps in alleviating the potential test data information leakage. 

\paragraph{SWAD consistently ranks 1st in every leaderboard. }
In leaderboards of supervised pretraining, self-supervised pretraining, and training from scratch, SWAD always ranks 1st and consistently improves upon ERM. This strongly demonstrates its effectiveness and universality in OOD generalization. 
Such a result, along with the high rank of RSC in the new leaderboards, indicates that intrinsic generalization properties and mechanisms like flatness or dropout could be less sensitive to test data information leakage and closer to essence of OOD generalization. 

In conclusion, we present the new leaderboards using the modified protocol to mitigate the possible test data information leakage, so that we can promote a fairer and more accurate evaluation and comparison between the domain generalization algorithms for their OOD generalization ability.

\section{Discussion}

\subsection{Position of pretraining for OOD generalization}
With the rapid development of pretrained models, including large language models like ChatGPT and LLaMA~\cite{touvron2023llama}, and large multi-modal models like GPT-4V~\cite{openai2023gpt4} and LLaVA~\cite{liu2023visual}, nowadays typically we tend to take advantage of pretraining in real applications. 
Despite their remarkable performance, current studies show that they are far from perfect and still suffer from performance degradation under distribution shifts, even for GPT-3.5~\cite{yang2022glue,yuan2023revisiting} and GPT-4V~\cite{zhang2024out}. Besides, it is hard to collect sufficient diverse data to train or fine-tune large models due to high expenses and privacy issues in some areas like medical care, where distribution shifts prevail. 
Thus the problem of OOD generalization still holds great significance in the era of large pretrained models. 

However, with the existence of pretrained models, proper evaluation of OOD generalization becomes a natural problem~\cite{yu2024survey}. 
Since we focus on the ability of models generalizing from training data to test data, strong pretrained weights naturally bring about possible test information leakage as we have analyzed in \cref{sec:pretrain}. 
\citet{kumar2021fine} also provide theoretical analyses that with good pretrained weights and a strong distribution shift, LP will outperform FT in that FT can distort the pretrained features, which are already good enough to generalize to test data. 
Our work serves as an initial effort to provide a better evaluation protocol for OOD generalization under pretraining. 

Recently, there have been works showing that pretraining on larger datasets with larger architecture backbones greatly improves test performance in OOD tasks \cite{kim2022broad, yu2021empirical, wiles2021fine}, and some directly design algorithms for better utilization of pretrained weights to improve test domain performance \cite{rame2023model, li2022simple}. 
We believe this is an interesting and meaningful research direction of valuable practical usage. If we directly focus on improving test performance, we should employ pretrained weights that are as strong and powerful as possible. However, for a fairer and more accurate evaluation of OOD generalization from training data to test data during the fine-tuning stage, we should seek pretrained weights that exhibit less test data information leakage.

\subsection{OOD model selection}
Model selection is another important topic for OOD generalization. DomainBed \cite{gulrajani2020search} has pointed out that oracle model selection (test-domain validation) leaks test data information in domain generalization. A similar problem also exists in subpopulation shift and is even more severe. \citet{idrissi2022simple} demonstrate that model selection based on the average accuracy on an IID validation set can lead to significant performance degradation compared with using worst group accuracy. The latter strategy utilizes the group label information, which should be considered as oracle model selection at least for methods claiming no need for group information. 
In this paper, we adhere to DomainBed's IID validation (training-domain validation) to prevent test information leakage for the guarantee of a fair and accurate comparison, but it does not mean that IID validation is the only right way for model selection. Considering the natural inconsistency of performance on IID validation data and test data~\cite{teney2023id}, maybe IID validation is not the best approach to guiding model selection. Improving the model selection strategy remains a fundamental research problem for OOD generalization in the future. 

\subsection{Datasets for domain generalization}

It has been a long time since domain generalization algorithms are primarily evaluated on relatively small datasets, e.g. Colored MNIST \cite{arjovsky2019invariant}, Rotated MNIST \cite{ghifary2015domain}, PACS \cite{li2017deeper}, VLCS \cite{khosla2012undoing}, OfficeHome \cite{venkateswara2017deep}, and TerraInc \cite{beery2018recognition}. 
Based on our analyses in this paper, our recommendation for expanding the number of test domains requires the establishment of larger datasets, and it is the same if we want to add more datasets to the leaderboard of training from scratch, which only DomainNet \cite{peng2019moment} and NICO++ \cite{zhang2023nico++} can support currently. 
It is much more challenging to construct a large, high-quality, and balanced dataset for domain generalization than to construct one for IID generalization, as it involves creating many domains with common classes. 
Recently, \citet{lynch2023spawrious} released a dataset Spawrious for better evaluation of OOD generalization, whose data is completely generated and collected from stable diffusion models. This provides a new direction for constructing domain generalization datasets with much lower cost. 

\section{Acknowledgements}

This work was supported in part by National Natural Science Foundation of China (No. 62141607), China National Postdoctoral Program for Innovative Talents (BX20230195). 
Peng Cui is the corresponding author. 
All opinions in this paper are those of the authors and do not necessarily reflect the views of the funding agencies.

{
    \small
    \bibliographystyle{ieeenat_fullname}
    \bibliography{main}
}

\appendix
\clearpage

\section{Related Work}
Given multiple training domains, the task of domain generalization focuses on improving the generalization ability of models on unseen test domains. Here we introduce related work of domain generalization from two perspectives: algorithms and evaluation. 
For other Out-of-Distribution (OOD) generalization algorithms, you may refer to the survey of OOD generalization~\cite{shen2021towards}. 
For a more detailed review of their evaluation, you may refer to the survey of OOD evaluation~\cite{yu2024survey}. 

\paragraph{Algorithms} There are many types of domain generalization algorithms. The most typical practice is learning representations that are invariant across training domains~\cite{mahajan2021domain,li2018domain,li2018domain2,li2018deep,xu2014exploiting,li2017domain, rame2022fishr}. Other common types of algorithms include data augmentation~\cite{shankar2018generalizing,volpi2018generalizing,yan2020improve}, meta learning~\cite{li2018learning,balaji2018metareg,li2019feature,li2019episodic}, sample reweighting~\cite{zhang2021deep}, and self-supervised learning~\cite{zhang2022udg,harary2022unsupervised}. Recently, more works focus on the fundamental properties and mechanisms of generalization, including emphasizing hard-to-learn features~\cite{huang2020self} or samples~\cite{huang2022two}, flatness~\cite{cha2021swad,zhang2023gradient,zhang2023flatness}, heterogeneity~\cite{liu2021heterogeneous,liu2022measure,tong2023quantitatively}, etc. 
For a more detailed review of domain generalization algorithms, you may refer to the surveys of domain generalization~\cite{wang2022generalizing, zhou2022domain}. 

\paragraph{Evaluation} To facilitate the appropriate evaluation of domain generalization algorithms, there are a number of datasets built with different types of distribution shifts. Among the most widely used ones, PACS~\cite{li2017deeper}, OfficeHome~\cite{venkateswara2017deep} and DomainNet~\cite{peng2019moment} depict the distribution shift via the change of image styles. VLCS~\cite{khosla2012undoing} is collected from multiple sources and treats each source as a domain. NICO++~\cite{zhang2023nico++,zhang2022nico} creates different contexts based on various elements like background, object color, season, etc. TerraInc~\cite{beery2018recognition} employs the camera view to generate distribution shifts. Apart from these datasets, \citet{gulrajani2020search} establishes DomainBed, a benchmark along with a comprehensive evaluation protocol by incorporating hyperparameter search into the evaluation pipeline, to bring convenience and fairness to the development and comparison of domain generalization algorithms.

\section{Additional experiments for the analyses of pretraining}
\label{supp:pretrain}

Note that for both linear-probing (LP) and fine-tuning (FT), we have searched the best hyperparameters on a validation set that shares the same distribution with the training domains, following the original protocol of DomainBed. 

\subsection{LP v.s. FT for supervised pretrained ResNet-50}
\label{supp:supervised}

Results of LP and FT with supervised pretrained ResNet-50 on DomainNet and NICO++ are shown in \cref{table:supp-supervised}. We can see that the phenomenon of LP outperforming FT still exists in some domains, which demonstrates the prevalence of such phenomenon under the standard domain generalization protocol. 

\subsection{LP v.s. FT for various pretraining}
\label{supp:various}

We change different pretraining architectures and methods. For PACS and VLCS, we conduct experiments with the same pretraining strategies and network backbones as in \cref{table:officehome}. For TerraInc, DomainNet, NICO++, we conduct experiments only with MoCo-v2 pretrained ResNet-50 and MoCo-v3 pretrained ViT-B/16. As shown in \cref{table:supp-pretrain} and \cref{table:supp-pretrain2}, we find that in these datasets, self-supervised pretraining still guarantees that linear-probing does not outperform fine-tuning, which is beneficial for a fairer comparison between algorithms.

\section{Detailed Leaderboards}
\label{supp:leaderboard}

Next, we present detailed leaderboards of each dataset. Our code is available at \url{https://github.com/h-yu16/DomainBed-v2}. Our code is mostly based on the implementation of DomainBed\footnote{\url{https://github.com/facebookresearch/DomainBed}\label{code:domainbed}}. 
Since most results on TerraInc fluctuate a lot, exhibiting a large standard deviation~\cite{gulrajani2020search}, we do not include this dataset in our leaderboards. 
Besides, we do not include Colored MNIST~\cite{arjovsky2019invariant} and Rotated MNIST~\cite{ghifary2015domain} since they are semi-synthetic and could be easily solved by data augmentation techniques like rotation.

\subsection{Supervised pretrained ResNet-50}

Results of PACS, VLCS, OfficeHome, and DomainNet in \cref{leaderboard} are mostly from the README of the official code repository of DomainBed\textsuperscript{\ref{code:domainbed}}, except the results of SWAD~\cite{cha2021swad} and Fishr~\cite{rame2022fishr} that are from their own paper. Note that in Fishr's main paper, only results of test-domain validation are presented, while we employ the results of training-domain validation in its appendix. Besides, detailed results of NICO++ are listed in \cref{supp:supervised-nico++}. 

\subsection{MoCo-v2 pretrained ResNet-50}
\label{supp:moco-v2}

Detailed results are listed from \cref{supp:mocov2-pacs} to \cref{supp:mocov2-nico}. 

\subsection{MoCo-v3 pretrained ViT-B/16}
\label{supp:mocov-v3}

Detailed results are listed from \cref{supp:mocov3-pacs} to \cref{supp:mocov3-nico}. 

\subsection{ResNet-50 trained from scratch}
\label{supp:none}

For the predefined hyperparameter search space, we adjust the space of learning rate from the logarithm of $10^{{\rm uniform}(-5,-3.5)}$ to that of $10^{{\rm uniform}(-4.5,-3)}$ because training from scratch requires a relatively higher learning rate. The number of iterations is set as 60,000. 
Detailed results are listed in \cref{supp:none-dn} and \cref{supp:none-nico}.

\begin{table*}[]
\centering
\caption{Results of linear-probing (LP) and fine-tuning (FT) with different pretraining methods of different backbone architectures on PACS, VLCS and TerraInc. }
\label{table:supp-pretrain}
\resizebox{0.75\textwidth}{!}{%
\begin{tabular}{@{}ccc|cccc|c@{}}
\toprule
\multicolumn{3}{c|}{PACS}                                                                                & P                 & A        & C        & S        & Avg      \\ \midrule
\multicolumn{1}{c|}{\multirow{2}{*}{ResNet-18}}  & \multicolumn{1}{c|}{\multirow{2}{*}{Supervised}} & LP & \textbf{95.8±0.2} & 69.9±0.7 & 52.5±0.7 & 48.7±1.4 & 66.7±0.5 \\
\multicolumn{1}{c|}{}                            & \multicolumn{1}{c|}{}                            & FT & 92.2±1.0          & 80.5±2.0 & 74.3±1.5 & 69.0±1.0 & 79.0±0.4 \\ \midrule
\multicolumn{1}{c|}{\multirow{10}{*}{ResNet-50}} & \multicolumn{1}{c|}{\multirow{2}{*}{Supervised}} & LP & \textbf{97.7±0.1} & 71.8±1.6 & 53.8±1.8 & 45.9±1.7 & 67.3±0.3 \\
\multicolumn{1}{c|}{}                            & \multicolumn{1}{c|}{}                            & FT & 97.4±0.1          & 86.1±0.9 & 80.4±1.4 & 77.1±2.5 & 85.3±0.6 \\ \cmidrule(l){2-8} 
\multicolumn{1}{c|}{}                            & \multicolumn{1}{c|}{\multirow{2}{*}{MoCo}}       & LP & 76.2±0.2          & 53.3±0.8 & 30.6±0.2 & 33.0±1.1 & 48.3±0.2 \\
\multicolumn{1}{c|}{}                            & \multicolumn{1}{c|}{}                            & FT & 92.4±0.4          & 75.3±2.4 & 71.0±1.8 & 69.0±2.5 & 76.9±1.4 \\ \cmidrule(l){2-8} 
\multicolumn{1}{c|}{}                            & \multicolumn{1}{c|}{\multirow{2}{*}{MoCo-v2}}    & LP & 91.9±0.0          & 70.0±0.3 & 51.0±0.1 & 32.7±0.9 & 61.4±0.2 \\
\multicolumn{1}{c|}{}                            & \multicolumn{1}{c|}{}                            & FT & 95.4±0.7          & 84.8±1.1 & 79.5±1.0 & 76.7±1.2 & 84.1±0.3 \\ \cmidrule(l){2-8} 
\multicolumn{1}{c|}{}                            & \multicolumn{1}{c|}{\multirow{2}{*}{SimCLR}}     & LP & 47.4±0.2          & 29.6±0.4 & 35.5±0.4 & 32.3±0.9 & 36.2±0.2 \\
\multicolumn{1}{c|}{}                            & \multicolumn{1}{c|}{}                            & FT & 73.4±1.6          & 52.0±2.1 & 60.9±2.0 & 57.8±1.8 & 61.0±0.2 \\ \cmidrule(l){2-8} 
\multicolumn{1}{c|}{}                            & \multicolumn{1}{c|}{\multirow{2}{*}{SimCLR-v2}}  & LP & 32.5±0.8          & 26.0±0.3 & 22.6±1.4 & 20.8±1.6 & 25.5±0.5 \\
\multicolumn{1}{c|}{}                            & \multicolumn{1}{c|}{}                            & FT & 89.7±1.5          & 75.7±1.6 & 75.5±2.3 & 71.8±2.1 & 78.2±1.6 \\ \midrule
\multicolumn{1}{c|}{\multirow{4}{*}{ViT-B/16}}   & \multicolumn{1}{c|}{\multirow{2}{*}{Supervised}} & LP & 88.6±2.6          & 76.3±0.4 & 64.9±0.6 & 48.2±1.7 & 69.5±0.7 \\
\multicolumn{1}{c|}{}                            & \multicolumn{1}{c|}{}                            & FT & 98.6±0.2          & 90.4±0.6 & 81.7±0.8 & 74.8±3.4 & 86.4±0.6 \\ \cmidrule(l){2-8} 
\multicolumn{1}{c|}{}                            & \multicolumn{1}{c|}{\multirow{2}{*}{MoCo-v3}}    & LP & 94.2±0.5          & 75.2±0.9 & 61.5±0.5 & 46.9±0.6 & 69.4±0.2 \\
\multicolumn{1}{c|}{}                            & \multicolumn{1}{c|}{}                            & FT & 97.3±0.2          & 87.9±1.6 & 83.0±0.4 & 75.0±0.8 & 85.8±0.3 \\ \bottomrule
\end{tabular}%
}

\vspace{10pt}

\resizebox{0.75\textwidth}{!}{%
\begin{tabular}{@{}ccc|cccc|c@{}}
\toprule
\multicolumn{3}{c|}{VLCS}                                                                       & V                 & L        & C                 & S                 & Avg               \\ \midrule
\multicolumn{1}{c|}{\multirow{2}{*}{ResNet-18}}  & \multicolumn{1}{c|}{\multirow{2}{*}{Supervised}} & LP & \textbf{76.0±0.7} & 58.6±0.1 & \textbf{97.8±0.5} & \textbf{70.1±0.4} & \textbf{75.6±0.0} \\
\multicolumn{1}{c|}{}                            & \multicolumn{1}{c|}{}                            & FT & 68.8±1.9          & 61.7±2.0 & 94.9±1.5          & 67.8±2.7          & 73.3±1.3          \\ \midrule
\multicolumn{1}{c|}{\multirow{10}{*}{ResNet-50}} & \multicolumn{1}{c|}{\multirow{2}{*}{Supervised}} & LP & \textbf{77.2±1.6} & 58.1±0.6 & \textbf{97.4±0.4} & 71.4±1.1          & 76.0±0.6          \\
\multicolumn{1}{c|}{}                            & \multicolumn{1}{c|}{}                            & FT & 73.5±1.5          & 66.3±0.9 & 96.9±1.1          & 71.7±1.5          & 77.1±1.0          \\ \cmidrule(l){2-8} 
\multicolumn{1}{c|}{}                            & \multicolumn{1}{c|}{\multirow{2}{*}{MoCo}}       & LP & 67.7±0.7          & 54.8±0.4 & 91.9±0.3          & 66.4±0.4          & 70.2±0.3          \\
\multicolumn{1}{c|}{}                            & \multicolumn{1}{c|}{}                            & FT & 69.6±1.1          & 63.6±0.7 & 96.4±2.1          & 69.4±1.2          & 74.8±0.7          \\ \cmidrule(l){2-8} 
\multicolumn{1}{c|}{}                            & \multicolumn{1}{c|}{\multirow{2}{*}{MoCo-v2}}    & LP & 71.9±0.1          & 54.5±0.4 & 97.6±0.2          & 70.8±0.9          & 73.7±0.2          \\
\multicolumn{1}{c|}{}                            & \multicolumn{1}{c|}{}                            & FT & 73.8±3.3          & 64.3±1.9 & 97.8±0.3          & 71.8±1.0          & 76.9±0.8          \\ \cmidrule(l){2-8} 
\multicolumn{1}{c|}{}                            & \multicolumn{1}{c|}{\multirow{2}{*}{SimCLR}}     & LP & 48.8±0.2          & 54.4±0.1 & 52.7±1.4          & 45.0±0.2          & 50.2±0.4          \\
\multicolumn{1}{c|}{}                            & \multicolumn{1}{c|}{}                            & FT & 56.3±1.2          & 62.8±0.3 & 75.3±2.9          & 58.9±1.1          & 63.3±0.4          \\ \cmidrule(l){2-8} 
\multicolumn{1}{c|}{}                            & \multicolumn{1}{c|}{\multirow{2}{*}{SimCLR-v2}}  & LP & 45.8±0.7          & 52.9±0.1 & 62.8±1.1          & 45.1±0.1          & 51.6±0.2          \\
\multicolumn{1}{c|}{}                            & \multicolumn{1}{c|}{}                            & FT & 63.8±1.2          & 63.8±1.6 & 86.8±2.7          & 65.0±2.3          & 69.9±1.1          \\ \midrule
\multicolumn{1}{c|}{\multirow{4}{*}{ViT-B/16}}   & \multicolumn{1}{c|}{\multirow{2}{*}{Supervised}} & LP & 66.0±0.2          & 62.4±0.8 & 91.7±0.8          & 72.0±0.5          & 73.0±0.4          \\
\multicolumn{1}{c|}{}                            & \multicolumn{1}{c|}{}                            & FT & 77.9±1.4          & 65.6±0.9 & 97.4±0.6          & 77.0±1.2          & 79.5±0.5          \\ \cmidrule(l){2-8} 
\multicolumn{1}{c|}{}                            & \multicolumn{1}{c|}{\multirow{2}{*}{MoCo-v3}}    & LP & 72.4±0.7          & 59.1±0.3 & 97.6±0.1          & 75.4±0.7          & 76.1±0.2          \\
\multicolumn{1}{c|}{}                            & \multicolumn{1}{c|}{}                            & FT & 74.4±1.6          & 64.9±0.5 & 98.2±0.4          & 76.1±0.2          & 78.4±0.6          \\ \bottomrule
\end{tabular}%
}

\vspace{10pt}

\resizebox{0.75\textwidth}{!}{%
\begin{tabular}{@{}cc|cccc|c@{}}
\toprule
\multicolumn{2}{c|}{TerraInc}                                & L38      & L43      & L46      & L100     & Avg      \\ \midrule
\multicolumn{1}{c|}{\multirow{2}{*}{ResNet-50 MoCo-v2}} & LP & 71.9±0.1 & 54.5±0.4 & 97.6±0.2 & 70.8±0.9 & 73.7±0.2 \\
\multicolumn{1}{c|}{}                                   & FT & 73.8±3.3 & 64.3±1.9 & 97.8±0.3 & 71.8±1.0 & 76.9±0.8 \\ \midrule
\multicolumn{1}{c|}{\multirow{2}{*}{ViT-B/16 MoCo-v3}}  & LP & 72.4±0.7 & 59.1±0.3 & 97.6±0.1 & 75.4±0.7 & 76.1±0.2 \\
\multicolumn{1}{c|}{}                                   & FT & 74.4±1.6 & 64.9±0.5 & 98.2±0.4 & 76.1±0.2 & 78.4±0.6 \\ \bottomrule
\end{tabular}%
}

\end{table*}

\begin{table*}[h]
\centering
\caption{Results of linear-probing (LP) and fine-tuning (FT) with supervised
pretrained ResNet-50 on DomainNet and NICO++. }
\label{table:supp-supervised}
\resizebox{0.9\textwidth}{!}{%
\begin{tabular}{@{}c|cccccc|c@{}}
\toprule
DomainNet & clipart  & infograph & painting & quickdraw         & real              & sketch   & Avg      \\ \midrule
LP        & 42.1±0.2 & 14.5±0.1  & 41.8±0.1 & 3.4±0.1           & \textbf{61.9±0.1} & 31.9±0.4 & 32.6±0.1 \\
FT        & 62.0±0.2 & 19.0±0.2  & 45.9±0.4 & 12.6±0.6          & 59.0±0.6          & 50.5±0.3 & 41.5±0.2 \\ \midrule
NICO++    & autumn   & rock      & dim      & grass             & outdoor           & water    & Avg      \\ \midrule
LP        & 77.1±0.1 & 76.6±0.1  & 66.1±0.0 & \textbf{80.7±0.2} & \textbf{75.6±0.1} & 68.3±0.3 & 74.1±0.0 \\
FT        & 80.0±0.9 & 78.6±0.2  & 72.5±0.7 & 80.1±0.4          & 75.3±0.1          & 70.0±0.3 & 76.1±0.1 \\ \bottomrule
\end{tabular}%
}
\end{table*}

\begin{table*}[]
\caption{Results of linear-probing (LP) and fine-tuning (FT) with different pretraining methods of different backbone architectures on DomainNet and NICO++. }
\label{table:supp-pretrain2}
\resizebox{\textwidth}{!}{%
\begin{tabular}{@{}cc|cccccl|l@{}}
\toprule
\multicolumn{2}{c|}{DomainNet}                               & clipart  & infograph & painting & quickdraw & real     & sketch   & Avg      \\ \midrule
\multicolumn{1}{c|}{\multirow{2}{*}{ResNet-50 MoCo-v2}} & LP & 16.7±0.1 & 07.3±0.1  & 28.4±0.2 & 0.6±0.0   & 33.6±0.1 & 18.4±0.1 & 17.5±0.0 \\
\multicolumn{1}{c|}{}                                   & FT & 59.4±0.6 & 17.9±0.4  & 45.4±0.6 & 12.9±0.6  & 57.0±0.6 & 49.8±0.4 & 40.4±0.3 \\ \midrule
\multicolumn{1}{c|}{\multirow{2}{*}{ViT-B/16 MoCo-v3}}  & LP & 29.3±0.2 & 13.9±0.1  & 39.3±0.2 & 2.2±0.1   & 43.2±0.2 & 27.5±0.1 & 25.9±0.1 \\
\multicolumn{1}{c|}{}                                   & FT & 68.9±0.2 & 24.7±0.4  & 55.5±1.1 & 17.0±1.1  & 67.6±0.2 & 56.3±0.2 & 48.3±0.1 \\ \bottomrule
\end{tabular}%
}

\vspace{10pt}

\resizebox{\textwidth}{!}{%
\begin{tabular}{@{}cc|cccccl|l@{}}
\toprule
\multicolumn{2}{c|}{NICO++}                                  & autumn   & rock     & dim      & grass    & outdoor  & \multicolumn{1}{c|}{water} & \multicolumn{1}{c}{Avg} \\ \midrule
\multicolumn{1}{c|}{\multirow{2}{*}{ResNet-50 MoCo-v2}} & LP & 64.5±0.3 & 61.8±0.3 & 47.2±0.0 & 64.7±0.1 & 59.1±0.2 & 50.5±0.1                   & 58.0±0.1                \\
\multicolumn{1}{c|}{}                                   & FT & 75.6±0.6 & 73.0±0.2 & 67.9±1.0 & 75.3±0.3 & 70.6±0.4 & 64.3±0.3                   & 71.1±0.1                \\ \midrule
\multicolumn{1}{c|}{\multirow{2}{*}{ViT-B/16 MoCo-v3}}  & LP & 78.1±0.2 & 77.2±0.1 & 62.8±0.2 & 80.3±0.1 & 73.1±0.2 & 65.7±0.1                   & 72.9±0.1                \\
\multicolumn{1}{c|}{}                                   & FT & 83.3±0.2 & 82.7±0.3 & 76.4±0.1 & 83.8±0.3 & 78.9±0.2 & 72.7±0.5                   & 79.6±0.2                \\ \bottomrule
\end{tabular}%
}

\end{table*}

\begin{table*}[]
\caption{Results of supervised pretrained ResNet-50 on NICO++. }
\label{supp:supervised-nico++}
\centering
\resizebox{0.9\textwidth}{!}{%
\begin{tabular}{@{}cccccccc@{}}
\toprule
\multicolumn{8}{c}{Supervised pretrained ResNet-50}                                                                             \\ \midrule
\multicolumn{1}{c|}{NICO++}   & autumn   & rock     & dim      & grass    & outdoor  & \multicolumn{1}{c|}{water}    & Avg      \\ \midrule
\multicolumn{1}{c|}{ERM}      & 81.0±0.3 & 78.8±0.4 & 73.7±0.4 & 80.5±0.4 & 78.8±0.2 & \multicolumn{1}{c|}{72.4±0.2} & 77.5±0.1 \\
\multicolumn{1}{c|}{SWAD}     & 83.3±0.2 & 81.8±0.1 & 76.6±0.2 & 83.3±0.1 & 80.8±0.1 & \multicolumn{1}{c|}{75.5±0.1} & 80.2±0.1 \\
\multicolumn{1}{c|}{RSC}      & 81.3±0.3 & 79.6±0.7 & 74.1±0.5 & 81.2±0.7 & 79.1±0.1 & \multicolumn{1}{c|}{73.2±0.3} & 78.1±0.2 \\
\multicolumn{1}{c|}{GroupDRO} & 80.9±0.7 & 79.3±0.3 & 73.4±0.7 & 80.4±0.2 & 78.9±0.6 & \multicolumn{1}{c|}{72.8±0.5} & 77.6±0.4 \\
\multicolumn{1}{c|}{Fishr}    & 81.6±0.3 & 80.4±0.1 & 74.0±0.5 & 81.9±0.5 & 79.6±0.5 & \multicolumn{1}{c|}{73.1±0.2} & 78.4±0.1 \\
\multicolumn{1}{c|}{CORAL}    & 82.5±0.6 & 81.3±0.1 & 74.4±0.2 & 82.4±0.2 & 80.3±0.2 & \multicolumn{1}{c|}{75.2±0.1} & 79.3±0.1 \\
\multicolumn{1}{c|}{MMD}      & 81.0±0.5 & 79.0±0.1 & 73.1±0.2 & 80.1±0.1 & 77.9±0.4 & \multicolumn{1}{c|}{72.8±1.5} & 77.3±0.2 \\
\multicolumn{1}{c|}{SagNet}   & 81.6±0.3 & 80.2±0.5 & 74.0±0.3 & 81.8±0.2 & 79.3±0.2 & \multicolumn{1}{c|}{74.4±0.2} & 78.6±0.2 \\
\multicolumn{1}{c|}{IRM}      & 80.3±1.3 & 79.0±0.2 & 73.2±0.9 & 81.2±0.4 & 78.6±0.1 & \multicolumn{1}{c|}{72.9±0.3} & 77.5±0.3 \\
\multicolumn{1}{c|}{Mixup}    & 82.4±0.1 & 80.6±0.2 & 74.5±0.4 & 81.5±0.3 & 80.0±0.3 & \multicolumn{1}{c|}{74.3±0.3} & 78.9±0.0 \\ \bottomrule
\end{tabular}%
}
\end{table*}

\clearpage

\begin{table*}[htb]
\caption{Results of MoCo-v2 pretrained ResNet-50 on PACS. }
\label{supp:mocov2-pacs}
\centering
\resizebox{0.7\textwidth}{!}{%
\begin{tabular}{@{}cccccc@{}}
\toprule
\multicolumn{6}{c}{MoCo-v2 pretrained ResNet-50}                                                          \\ \midrule
\multicolumn{1}{c|}{PACS}     & photo    & art      & cartoon  & \multicolumn{1}{c|}{sketch}   & Avg      \\ \midrule
\multicolumn{1}{c|}{ERM}      & 95.4±0.7 & 84.8±1.1 & 79.5±1.0 & \multicolumn{1}{c|}{76.7±1.2} & 84.1±0.3 \\
\multicolumn{1}{c|}{SWAD}     & 96.8±0.3 & 88.1±0.5 & 82.4±0.2 & \multicolumn{1}{c|}{77.4±1.8} & 86.2±0.6 \\
\multicolumn{1}{c|}{RSC}      & 97.6±0.7 & 87.9±1.3 & 81.7±0.7 & \multicolumn{1}{c|}{76.7±5.0} & 85.9±1.9 \\
\multicolumn{1}{c|}{GroupDRO} & 95.4±0.2 & 83.3±0.7 & 80.2±1.1 & \multicolumn{1}{c|}{73.2±1.3} & 83.0±0.5 \\
\multicolumn{1}{c|}{Fishr}    & 95.0±0.1 & 82.9±1.0 & 79.1±3.5 & \multicolumn{1}{c|}{74.1±0.5} & 82.8±1.0 \\
\multicolumn{1}{c|}{CORAL}    & 96.2±0.3 & 85.3±2.8 & 77.8±1.1 & \multicolumn{1}{c|}{76.7±1.7} & 84.0±1.4 \\
\multicolumn{1}{c|}{MMD}      & 95.4±0.5 & 85.8±1.2 & 80.6±0.8 & \multicolumn{1}{c|}{75.5±3.7} & 84.3±1.2 \\
\multicolumn{1}{c|}{SagNet}   & 92.6±0.5 & 83.7±2.3 & 78.2±0.5 & \multicolumn{1}{c|}{75.7±2.3} & 82.5±1.1 \\
\multicolumn{1}{c|}{IRM}      & 95.3±1.0 & 82.6±2.2 & 80.7±1.1 & \multicolumn{1}{c|}{75.6±5.3} & 83.5±0.8 \\
\multicolumn{1}{c|}{Mixup}    & 95.7±0.2 & 84.3±1.3 & 79.7±0.8 & \multicolumn{1}{c|}{70.4±4.6} & 82.5±1.6 \\ \bottomrule
\end{tabular}%
}
\end{table*}

\begin{table*}[htb]
\caption{Results of MoCo-v2 pretrained ResNet-50 on VLCS. }
\label{supp:mocov2-vlcs}
\centering
\resizebox{0.7\textwidth}{!}{%
\begin{tabular}{@{}cccccc@{}}
\toprule
\multicolumn{6}{c}{MoCo-v2 pretrained ResNet-50}                                                          \\ \midrule
\multicolumn{1}{c|}{VLCS}     & PASCAL   & LABELME  & CALTECH  & \multicolumn{1}{c|}{SUN}      & Avg      \\ \midrule
\multicolumn{1}{c|}{ERM}      & 73.8±3.3 & 64.3±1.9 & 97.8±0.3 & \multicolumn{1}{c|}{71.8±1.0} & 76.9±0.8 \\
\multicolumn{1}{c|}{SWAD}     & 75.7±1.0 & 61.9±0.9 & 98.5±0.3 & \multicolumn{1}{c|}{74.8±1.2} & 77.7±0.7 \\
\multicolumn{1}{c|}{RSC}      & 76.4±2.1 & 66.3±1.0 & 99.1±1.3 & \multicolumn{1}{c|}{72.2±0.5} & 78.5±0.3 \\
\multicolumn{1}{c|}{GroupDRO} & 72.8±2.2 & 64.7±0.4 & 96.2±0.9 & \multicolumn{1}{c|}{72.2±0.2} & 76.5±0.8 \\
\multicolumn{1}{c|}{Fishr}    & 72.6±2.9 & 62.7±1.9 & 96.0±0.8 & \multicolumn{1}{c|}{70.2±2.5} & 75.4±1.6 \\
\multicolumn{1}{c|}{CORAL}    & 74.7±1.5 & 64.7±1.0 & 97.3±0.8 & \multicolumn{1}{c|}{73.3±1.5} & 77.5±0.4 \\
\multicolumn{1}{c|}{MMD}      & 73.6±2.2 & 63.3±1.7 & 97.9±0.3 & \multicolumn{1}{c|}{72.1±2.5} & 76.7±1.3 \\
\multicolumn{1}{c|}{SagNet}   & 72.8±1.7 & 62.9±1.4 & 96.3±0.7 & \multicolumn{1}{c|}{70.1±1.1} & 75.5±0.8 \\
\multicolumn{1}{c|}{IRM}      & 70.0±1.5 & 62.6±0.8 & 98.1±0.9 & \multicolumn{1}{c|}{70.7±0.5} & 75.3±0.5 \\
\multicolumn{1}{c|}{Mixup}    & 71.4±1.2 & 63.2±2.5 & 97.8±0.2 & \multicolumn{1}{c|}{71.8±0.2} & 76.1±0.7 \\ \bottomrule
\end{tabular}%
}
\end{table*}

\begin{table*}[htb]
\caption{Results of MoCo-v2 pretrained ResNet-50 on OfficeHome. }
\label{supp:mocov2-oh}
\centering
\resizebox{0.7\textwidth}{!}{%
\begin{tabular}{@{}cccccc@{}}
\toprule
\multicolumn{6}{c}{MoCo-v2 pretrained ResNet-50}                                                            \\ \midrule
\multicolumn{1}{c|}{OfficeHome} & Art      & Clipart  & Product  & \multicolumn{1}{c|}{Real}     & Avg      \\ \midrule
\multicolumn{1}{c|}{ERM}        & 49.6±2.7 & 45.2±2.0 & 65.8±1.3 & \multicolumn{1}{c|}{68.6±0.2} & 57.3±0.3 \\
\multicolumn{1}{c|}{SWAD}       & 57.4±0.2 & 49.9±0.8 & 69.6±0.3 & \multicolumn{1}{c|}{73.6±0.4} & 62.6±0.1 \\
\multicolumn{1}{c|}{RSC}        & 54.0±0.9 & 46.4±1.3 & 69.1±1.0 & \multicolumn{1}{c|}{71.2±0.5} & 60.2±0.3 \\
\multicolumn{1}{c|}{GroupDRO}   & 50.0±1.4 & 45.5±1.1 & 66.2±0.8 & \multicolumn{1}{c|}{68.0±0.5} & 57.4±0.8 \\
\multicolumn{1}{c|}{Fishr}      & 52.5±1.4 & 45.7±0.7 & 67.3±0.6 & \multicolumn{1}{c|}{69.2±0.8} & 58.7±0.2 \\
\multicolumn{1}{c|}{CORAL}      & 58.2±0.5 & 49.2±0.4 & 71.1±0.5 & \multicolumn{1}{c|}{72.1±0.4} & 62.6±0.3 \\
\multicolumn{1}{c|}{MMD}        & 50.7±0.7 & 44.6±2.3 & 66.6±0.8 & \multicolumn{1}{c|}{68.4±0.6} & 57.6±0.8 \\
\multicolumn{1}{c|}{SagNet}     & 55.6±1.5 & 47.1±2.9 & 69.9±0.5 & \multicolumn{1}{c|}{72.1±0.4} & 61.2±1.0 \\
\multicolumn{1}{c|}{IRM}        & 50.1±1.1 & 45.2±1.4 & 67.0±1.2 & \multicolumn{1}{c|}{69.2±0.5} & 57.9±0.3 \\
\multicolumn{1}{c|}{Mixup}      & 54.7±0.8 & 47.5±1.2 & 67.8±0.5 & \multicolumn{1}{c|}{72.0±0.4} & 60.5±0.2 \\ \bottomrule
\end{tabular}%
}
\end{table*}

\begin{table*}[htb]
\caption{Results of MoCo-v2 pretrained ResNet-50 on DomainNet. }
\label{supp:mocov2-dn}
\centering
\resizebox{0.9\textwidth}{!}{%
\begin{tabular}{@{}cccccccc@{}}
\toprule
\multicolumn{8}{c}{MoCo-v2 pretrained ResNet-50}                                                                                    \\ \midrule
\multicolumn{1}{c|}{DomainNet} & clipart  & infograph & painting & quickdraw & real     & \multicolumn{1}{c|}{sketch}   & Avg      \\ \midrule
\multicolumn{1}{c|}{ERM}       & 57.8±0.6 & 17.5±0.4  & 46.5±0.7 & 12.7±0.8  & 54.8±0.2 & \multicolumn{1}{c|}{46.6±0.4} & 39.3±0.3 \\
\multicolumn{1}{c|}{SWAD}      & 62.1±0.1 & 19.1±0.1  & 50.5±0.0 & 15.0±0.2  & 59.4±0.2 & \multicolumn{1}{c|}{51.4±0.2} & 42.9±0.1 \\
\multicolumn{1}{c|}{RSC}       & 57.7±0.9 & 19.0±0.7  & 47.2±0.2 & 15.8±0.7  & 54.8±0.7 & \multicolumn{1}{c|}{48.5±0.4} & 40.5±0.4 \\
\multicolumn{1}{c|}{GroupDRO}  & 55.9±1.3 & 16.1±1.0  & 41.8±0.6 & 11.3±0.6  & 50.5±0.6 & \multicolumn{1}{c|}{42.3±0.4} & 36.3±0.4 \\
\multicolumn{1}{c|}{Fishr}     & 57.4±0.5 & 16.9±0.6  & 46.2±0.3 & 12.5±0.5  & 55.1±0.5 & \multicolumn{1}{c|}{47.7±0.1} & 39.3±0.3 \\
\multicolumn{1}{c|}{CORAL}     & 59.5±0.2 & 18.5±0.7  & 49.2±1.1 & 13.8±0.7  & 57.1±0.1 & \multicolumn{1}{c|}{48.8±0.7} & 41.2±0.1 \\
\multicolumn{1}{c|}{MMD}       & 58.0±0.4 & 17.2±0.1  & 46.4±0.9 & 13.1±0.3  & 54.4±0.9 & \multicolumn{1}{c|}{46.8±0.4} & 39.3±0.3 \\
\multicolumn{1}{c|}{SagNet}    & 57.4±0.2 & 16.0±0.4  & 46.1±0.6 & 12.2±0.8  & 53.2±0.3 & \multicolumn{1}{c|}{46.7±0.3} & 38.6±0.2 \\
\multicolumn{1}{c|}{IRM}       & 57.3±0.2 & 17.2±0.3  & 46.3±0.4 & 13.1±0.3  & 54.1±0.6 & \multicolumn{1}{c|}{46.7±0.4} & 39.1±0.1 \\
\multicolumn{1}{c|}{Mixup}     & 56.2±0.9 & 16.8±0.2  & 46.7±0.1 & 13.0±0.7  & 54.8±0.7 & \multicolumn{1}{c|}{47.8±0.3} & 39.2±0.1 \\ \bottomrule
\end{tabular}%
}
\end{table*}

\begin{table*}[htb]
\caption{Results of MoCo-v2 pretrained ResNet-50 on NICO++. }
\label{supp:mocov2-nico}
\centering
\resizebox{0.9\textwidth}{!}{%
\begin{tabular}{@{}cccccccc@{}}
\toprule
\multicolumn{8}{c}{MoCo-v2 pretrained ResNet-50}                                                                                 \\ \midrule
\multicolumn{1}{c|}{NICO++}   & autumn   & rock     & dim      & grass    & outdoor  & \multicolumn{1}{c|}{water}    & Avg      \\ \midrule
\multicolumn{1}{c|}{ERM}      & 75.6±0.6 & 73.0±0.2 & 67.9±1.0 & 75.3±0.3 & 70.6±0.4 & \multicolumn{1}{c|}{64.3±0.3} & 71.1±0.1 \\
\multicolumn{1}{c|}{SWAD}     & 80.2±0.1 & 78.2±0.1 & 72.7±0.1 & 80.1±0.3 & 74.8±0.1 & \multicolumn{1}{c|}{67.8±0.1} & 75.6±0.1 \\
\multicolumn{1}{c|}{RSC}      & 78.8±0.6 & 77.2±0.7 & 70.7±0.8 & 78.4±0.6 & 73.7±0.8 & \multicolumn{1}{c|}{66.5±0.9} & 74.2±0.5 \\
\multicolumn{1}{c|}{GroupDRO} & 76.4±0.8 & 74.3±0.4 & 67.4±1.0 & 75.2±0.5 & 70.2±0.2 & \multicolumn{1}{c|}{63.9±0.9} & 71.2±0.2 \\
\multicolumn{1}{c|}{Fishr}    & 75.1±0.9 & 73.7±1.1 & 67.1±0.1 & 75.1±0.7 & 70.7±0.5 & \multicolumn{1}{c|}{64.0±1.0} & 71.0±0.1 \\
\multicolumn{1}{c|}{CORAL}    & 78.2±0.5 & 76.7±0.6 & 69.2±0.3 & 78.2±0.3 & 73.1±0.6 & \multicolumn{1}{c|}{66.5±0.9} & 73.7±0.2 \\
\multicolumn{1}{c|}{MMD}      & 76.0±0.5 & 74.2±0.3 & 67.6±0.1 & 75.6±1.1 & 71.1±0.1 & \multicolumn{1}{c|}{64.8±0.6} & 71.6±0.3 \\
\multicolumn{1}{c|}{SagNet}   & 74.6±0.8 & 72.4±0.6 & 66.7±0.9 & 73.9±0.5 & 70.5±0.2 & \multicolumn{1}{c|}{63.1±0.7} & 70.2±0.3 \\
\multicolumn{1}{c|}{IRM}      & 76.0±1.0 & 73.9±0.7 & 68.1±0.4 & 75.8±0.4 & 71.6±0.4 & \multicolumn{1}{c|}{63.8±0.8} & 71.5±0.3 \\
\multicolumn{1}{c|}{Mixup}    & 77.4±0.2 & 76.5±0.1 & 68.8±0.7 & 77.6±0.5 & 73.1±0.3 & \multicolumn{1}{c|}{66.5±0.3} & 73.3±0.2 \\ \bottomrule
\end{tabular}%
}
\end{table*}

\clearpage

\begin{table*}[htb]
\caption{Results of MoCo-v3 pretrained ViT-B/16 on PACS. }
\label{supp:mocov3-pacs}
\centering
\resizebox{0.7\textwidth}{!}{%
\begin{tabular}{@{}cccccc@{}}
\toprule
\multicolumn{6}{c}{MoCo-v3 pretrained ViT-B/16}                                                           \\ \midrule
\multicolumn{1}{c|}{PACS}     & photo    & art      & cartoon  & \multicolumn{1}{c|}{sketch}   & Avg      \\ \midrule
\multicolumn{1}{c|}{ERM}      & 97.3±0.2 & 87.9±1.6 & 83.0±0.4 & \multicolumn{1}{c|}{75.0±0.8} & 85.8±0.3 \\
\multicolumn{1}{c|}{SWAD}     & 98.4±0.4 & 91.9±0.0 & 83.8±0.5 & \multicolumn{1}{c|}{78.2±0.5} & 88.1±0.3 \\
\multicolumn{1}{c|}{RSC}      & 98.3±0.5 & 89.2±1.2 & 83.8±1.5 & \multicolumn{1}{c|}{75.8±1.7} & 86.8±0.5 \\
\multicolumn{1}{c|}{GroupDRO} & 97.6±0.2 & 89.5±2.2 & 82.8±0.3 & \multicolumn{1}{c|}{72.5±2.2} & 85.6±0.8 \\
\multicolumn{1}{c|}{Fishr}    & 97.7±0.2 & 88.1±0.9 & 85.0±1.7 & \multicolumn{1}{c|}{75.7±3.7} & 86.6±1.3 \\
\multicolumn{1}{c|}{CORAL}    & 97.8±0.3 & 84.4±2.7 & 83.9±1.1 & \multicolumn{1}{c|}{80.5±1.4} & 86.7±0.9 \\
\multicolumn{1}{c|}{MMD}      & 98.7±0.0 & 87.6±1.6 & 83.7±0.4 & \multicolumn{1}{c|}{75.7±2.4} & 86.4±0.9 \\
\multicolumn{1}{c|}{SagNet}   & 97.0±0.4 & 88.3±0.9 & 82.9±2.9 & \multicolumn{1}{c|}{73.5±3.0} & 85.6±1.0 \\
\multicolumn{1}{c|}{IRM}      & 96.9±0.0 & 87.7±1.1 & 82.7±1.7 & \multicolumn{1}{c|}{71.6±0.3} & 84.7±0.5 \\
\multicolumn{1}{c|}{Mixup}    & 97.3±0.4 & 87.1±0.8 & 84.4±0.9 & \multicolumn{1}{c|}{64.7±2.8} & 83.4±0.8 \\ \bottomrule
\end{tabular}%
}
\end{table*}

\begin{table*}[htb]
\caption{Results of MoCo-v3 pretrained ViT-B/16 on VLCS. }
\label{supp:mocov3-vlcs}
\centering
\resizebox{0.7\textwidth}{!}{%
\begin{tabular}{@{}cccccc@{}}
\toprule
\multicolumn{6}{c}{MoCo-v3 pretrained ViT-B/16}                                                            \\ \midrule
\multicolumn{1}{c|}{VLCS}     & PASCAL   & LABELME  & CALTECH  & \multicolumn{1}{c|}{SUN}       & Avg      \\ \midrule
\multicolumn{1}{c|}{ERM}      & 74.4±1.6 & 64.9±0.5 & 98.2±0.4 & \multicolumn{1}{c|}{76.1±0.2}  & 78.4±0.6 \\
\multicolumn{1}{c|}{SWAD}     & 77.4±0.5 & 64.0±0.4 & 97.6±0.6 & \multicolumn{1}{c|}{77.1±0.4}  & 79.0±0.1 \\
\multicolumn{1}{c|}{RSC}      & 74.8±3.6 & 64.4±0.4 & 98.2±0.4 & \multicolumn{1}{c|}{75.6±0.5}  & 78.2±1.1 \\
\multicolumn{1}{c|}{GroupDRO} & 73.2±0.3 & 64.3±1.9 & 97.5±0.5 & \multicolumn{1}{c|}{77.9±1.6}  & 78.2±0.9 \\
\multicolumn{1}{c|}{Fishr}    & 74.1±1.2 & 64.4±1.3 & 98.0±0.6 & \multicolumn{1}{c|}{75.4±1.0}  & 78.0±0.4 \\
\multicolumn{1}{c|}{CORAL}    & 74.9±1.1 & 64.0±1.4 & 98.3±0.4 & \multicolumn{1}{c|}{75.2±1.9}  & 78.1±0.6 \\
\multicolumn{1}{c|}{MMD}      & 67.7±0.7 & 65.6±0.7 & 61.5±0.0 & \multicolumn{1}{c|}{63.1±17.5} & 64.4±4.3 \\
\multicolumn{1}{c|}{SagNet}   & 74.3±1.6 & 64.5±1.1 & 97.6±0.3 & \multicolumn{1}{c|}{75.7±0.4}  & 78.0±0.5 \\
\multicolumn{1}{c|}{IRM}      & 74.1±1.0 & 65.3±1.6 & 97.0±0.5 & \multicolumn{1}{c|}{76.1±0.5}  & 78.1±0.3 \\
\multicolumn{1}{c|}{Mixup}    & 74.4±2.2 & 65.1±0.9 & 98.0±0.4 & \multicolumn{1}{c|}{75.4±1.2}  & 78.2±0.8 \\ \bottomrule
\end{tabular}%
}
\end{table*}

\begin{table*}[htb]
\caption{Results of MoCo-v3 pretrained ViT-B/16 on OfficeHome. }
\label{supp:mocov3-oh}
\centering
\resizebox{0.7\textwidth}{!}{%
\begin{tabular}{@{}cccccc@{}}
\toprule
\multicolumn{6}{c}{MoCo-v3 pretrained ViT-B/16}                                                             \\ \midrule
\multicolumn{1}{c|}{OfficeHome} & Art      & Clipart  & Product  & \multicolumn{1}{c|}{Real}     & Avg      \\ \midrule
\multicolumn{1}{c|}{ERM}        & 64.7±0.6 & 54.1±0.5 & 74.8±0.7 & \multicolumn{1}{c|}{78.3±0.7} & 68.0±0.4 \\
\multicolumn{1}{c|}{SWAD}       & 71.0±0.5 & 54.5±1.0 & 78.7±0.2 & \multicolumn{1}{c|}{81.4±0.2} & 71.4±0.4 \\
\multicolumn{1}{c|}{RSC}        & 64.2±0.5 & 55.8±0.3 & 74.3±0.2 & \multicolumn{1}{c|}{77.3±0.5} & 67.9±0.1 \\
\multicolumn{1}{c|}{GroupDRO}   & 63.8±0.6 & 55.6±0.8 & 74.8±0.4 & \multicolumn{1}{c|}{77.9±0.6} & 68.0±0.2 \\
\multicolumn{1}{c|}{Fishr}      & 64.3±0.3 & 54.8±0.5 & 74.7±0.8 & \multicolumn{1}{c|}{77.1±0.8} & 67.7±0.3 \\
\multicolumn{1}{c|}{CORAL}      & 63.9±1.1 & 53.7±0.4 & 74.9±0.8 & \multicolumn{1}{c|}{78.6±0.4} & 67.8±0.5 \\
\multicolumn{1}{c|}{MMD}        & 63.6±0.5 & 53.6±0.9 & 74.4±0.2 & \multicolumn{1}{c|}{77.2±0.3} & 67.2±0.2 \\
\multicolumn{1}{c|}{SagNet}     & 63.1±1.2 & 52.9±0.9 & 73.9±0.4 & \multicolumn{1}{c|}{77.5±0.2} & 66.9±0.2 \\
\multicolumn{1}{c|}{IRM}        & 64.7±1.0 & 55.8±1.3 & 74.4±0.5 & \multicolumn{1}{c|}{77.9±0.3} & 68.2±0.5 \\
\multicolumn{1}{c|}{Mixup}      & 66.3±0.6 & 57.6±0.1 & 76.7±0.6 & \multicolumn{1}{c|}{79.3±0.3} & 70.0±0.1 \\ \bottomrule
\end{tabular}%
}
\end{table*}

\begin{table*}[htb]
\caption{Results of MoCo-v3 pretrained ViT-B/16 on DomainNet. }
\label{supp:mocov3-dn}
\centering
\resizebox{0.9\textwidth}{!}{%
\begin{tabular}{@{}cccccccc@{}}
\toprule
\multicolumn{8}{c}{MoCo-v3 pretrained ViT-B/16}                                                                                    \\ \midrule
\multicolumn{1}{c|}{DomainNet} & clipart  & infograph & painting & quickdraw & real     & \multicolumn{1}{c|}{sketch}   & Avg      \\ \midrule
\multicolumn{1}{c|}{ERM}       & 68.2±0.1 & 23.0±0.1  & 55.3±0.3 & 16.8±1.1  & 66.2±0.2 & \multicolumn{1}{c|}{54.8±0.4} & 47.4±0.2 \\
\multicolumn{1}{c|}{SWAD}      & 69.7±0.2 & 24.3±0.1  & 58.8±0.1 & 19.3±0.3  & 68.1±0.2 & \multicolumn{1}{c|}{57.4±0.3} & 49.6±0.1 \\
\multicolumn{1}{c|}{RSC}       & 68.3±0.1 & 22.8±0.2  & 55.6±0.4 & 17.0±0.8  & 65.7±0.2 & \multicolumn{1}{c|}{54.2±0.3} & 47.3±0.1 \\
\multicolumn{1}{c|}{GroupDRO}  & 66.2±0.5 & 22.0±0.2  & 52.9±0.4 & 16.1±1.2  & 62.0±0.3 & \multicolumn{1}{c|}{50.4±0.4} & 44.9±0.1 \\
\multicolumn{1}{c|}{Fishr}     & 68.3±0.2 & 23.2±0.1  & 55.5±0.4 & 16.8±0.6  & 65.7±0.1 & \multicolumn{1}{c|}{53.8±0.2} & 47.2±0.1 \\
\multicolumn{1}{c|}{CORAL}     & 68.4±0.2 & 23.2±0.2  & 55.3±0.5 & 17.4±0.8  & 66.3±0.2 & \multicolumn{1}{c|}{54.8±0.1} & 47.5±0.1 \\
\multicolumn{1}{c|}{MMD}       & 68.2±0.2 & 22.7±0.1  & 55.3±0.4 & 17.0±0.8  & 66.1±0.1 & \multicolumn{1}{c|}{54.5±0.2} & 47.3±0.1 \\
\multicolumn{1}{c|}{SagNet}    & 67.3±0.5 & 21.9±0.3  & 54.5±0.7 & 16.6±0.4  & 64.8±0.4 & \multicolumn{1}{c|}{54.0±0.5} & 46.5±0.1 \\
\multicolumn{1}{c|}{IRM}       & 68.1±0.2 & 22.3±0.2  & 55.3±0.3 & 17.5±0.7  & 66.0±0.2 & \multicolumn{1}{c|}{54.5±0.2} & 47.3±0.1 \\
\multicolumn{1}{c|}{Mixup}     & 67.8±0.4 & 23.9±0.3  & 57.3±0.5 & 17.3±0.5  & 65.9±0.3 & \multicolumn{1}{c|}{56.1±0.3} & 48.0±0.1 \\ \bottomrule
\end{tabular}%
}
\end{table*}

\begin{table*}[htb]
\caption{Results of MoCo-v3 pretrained ViT-B/16 on NICO++. }
\label{supp:mocov3-nico}
\centering
\resizebox{0.9\textwidth}{!}{%
\begin{tabular}{@{}cccccccc@{}}
\toprule
\multicolumn{8}{c}{MoCo-v3 pretrained ViT-B/16}                                                                                 \\ \midrule
\multicolumn{1}{c|}{NICO++}   & autumn   & rock     & dim      & grass    & outdoor  & \multicolumn{1}{c|}{water}    & Avg      \\ \midrule
\multicolumn{1}{c|}{ERM}      & 83.3±0.2 & 82.7±0.3 & 76.4±0.1 & 83.8±0.3 & 78.9±0.2 & \multicolumn{1}{c|}{72.7±0.5} & 79.6±0.2 \\
\multicolumn{1}{c|}{SWAD}     & 84.6±0.1 & 84.3±0.1 & 77.6±0.0 & 84.5±0.1 & 79.9±0.0 & \multicolumn{1}{c|}{74.0±0.1} & 80.8±0.0 \\
\multicolumn{1}{c|}{RSC}      & 83.1±0.3 & 83.0±0.3 & 76.0±0.1 & 84.0±0.4 & 79.2±0.4 & \multicolumn{1}{c|}{72.8±0.8} & 79.7±0.1 \\
\multicolumn{1}{c|}{GroupDRO} & 83.0±0.2 & 83.2±0.2 & 76.0±0.4 & 84.1±0.1 & 79.1±0.3 & \multicolumn{1}{c|}{72.9±0.6} & 79.7±0.2 \\
\multicolumn{1}{c|}{Fishr}    & 83.0±0.6 & 83.0±0.4 & 76.2±0.2 & 84.1±0.1 & 78.9±0.4 & \multicolumn{1}{c|}{72.6±0.4} & 79.6±0.2 \\
\multicolumn{1}{c|}{CORAL}    & 83.0±0.3 & 82.9±0.2 & 75.9±0.1 & 83.5±0.4 & 78.7±0.2 & \multicolumn{1}{c|}{73.1±0.3} & 79.5±0.1 \\
\multicolumn{1}{c|}{MMD}      & 73.6±1.1 & 72.6±0.9 & 65.3±0.0 & 75.5±0.3 & 69.8±1.9 & \multicolumn{1}{c|}{61.4±1.9} & 69.7±1.0 \\
\multicolumn{1}{c|}{SagNet}   & 83.0±0.4 & 82.7±0.3 & 75.4±0.7 & 83.7±0.1 & 78.3±0.3 & \multicolumn{1}{c|}{72.0±0.2} & 79.2±0.2 \\
\multicolumn{1}{c|}{IRM}      & 83.7±0.2 & 82.9±0.1 & 76.1±0.1 & 83.9±0.2 & 78.8±0.6 & \multicolumn{1}{c|}{72.9±0.3} & 79.7±0.1 \\
\multicolumn{1}{c|}{Mixup}    & 82.8±0.2 & 82.7±0.4 & 77.2±0.2 & 84.2±0.2 & 79.1±0.5 & \multicolumn{1}{c|}{73.0±0.4} & 79.8±0.2 \\ \bottomrule
\end{tabular}%
}
\end{table*}

\clearpage

\begin{table*}[htb]
\caption{Results of ResNet-50 trained from scratch on DomainNet. }
\label{supp:none-dn}
\centering
\resizebox{0.9\textwidth}{!}{%
\begin{tabular}{@{}cccccccc@{}}
\toprule
\multicolumn{8}{c}{ResNet-50 trained from scratch}                                                                                 \\ \midrule
\multicolumn{1}{c|}{DomainNet} & clipart  & infograph & painting & quickdraw & real     & \multicolumn{1}{c|}{sketch}   & Avg      \\ \midrule
\multicolumn{1}{c|}{ERM}       & 53.6±0.3 & 12.5±0.0  & 37.2±0.1 & 12.8±0.5  & 42.9±0.2 & \multicolumn{1}{c|}{40.8±0.7} & 33.3±0.1 \\
\multicolumn{1}{c|}{SWAD}      & 58.4±0.1 & 15.6±0.1  & 42.4±0.2 & 15.8±0.1  & 47.6±0.3 & \multicolumn{1}{c|}{45.3±0.3} & 37.5±0.1 \\
\multicolumn{1}{c|}{RSC}       & 54.3±0.4 & 13.5±0.3  & 38.6±0.3 & 14.5±0.1  & 43.6±0.2 & \multicolumn{1}{c|}{41.5±0.6} & 34.3±0.1 \\
\multicolumn{1}{c|}{GroupDRO}  & 51.9±0.4 & 11.0±0.4  & 34.1±0.5 & 11.4±0.7  & 32.7±0.8 & \multicolumn{1}{c|}{34.5±1.0} & 29.3±0.2 \\
\multicolumn{1}{c|}{Fishr}     & 53.6±0.5 & 12.6±0.2  & 37.6±0.4 & 13.3±1.0  & 43.4±0.1 & \multicolumn{1}{c|}{40.4±0.6} & 33.5±0.3 \\
\multicolumn{1}{c|}{CORAL}     & 53.7±0.4 & 11.5±0.3  & 37.2±0.5 & 11.3±0.7  & 36.4±0.3 & \multicolumn{1}{c|}{37.6±0.4} & 31.3±0.2 \\
\multicolumn{1}{c|}{MMD}       & 53.2±0.4 & 11.2±0.2  & 37.4±0.1 & 12.3±1.2  & 37.4±0.1 & \multicolumn{1}{c|}{38.7±0.6} & 31.7±0.2 \\
\multicolumn{1}{c|}{SagNet}    & 53.2±0.1 & 11.3±0.1  & 36.5±0.1 & 11.9±0.3  & 41.7±0.3 & \multicolumn{1}{c|}{38.6±0.5} & 32.2±0.1 \\
\multicolumn{1}{c|}{IRM}       & 53.2±0.9 & 12.2±0.1  & 37.5±0.4 & 13.8±0.4  & 42.7±0.3 & \multicolumn{1}{c|}{40.5±0.4} & 33.3±0.4 \\
\multicolumn{1}{c|}{Mixup}     & 54.3±0.9 & 12.8±0.4  & 41.3±0.7 & 12.7±0.3  & 42.5±0.8 & \multicolumn{1}{c|}{43.5±0.9} & 34.5±0.3 \\ \bottomrule
\end{tabular}%
}
\end{table*}

\begin{table*}[htb]
\caption{Results of ResNet-50 trained from scratch on NICO++. }
\label{supp:none-nico}
\centering
\resizebox{0.9\textwidth}{!}{%
\begin{tabular}{@{}cccccccc@{}}
\toprule
\multicolumn{8}{c}{ResNet-50 trained from scratch}                                                                              \\ \midrule
\multicolumn{1}{c|}{NICO++}   & autumn   & rock     & dim      & grass    & outdoor  & \multicolumn{1}{c|}{water}    & Avg      \\ \midrule
\multicolumn{1}{c|}{ERM}      & 54.8±0.4 & 51.8±0.4 & 47.1±0.1 & 51.3±0.3 & 48.5±0.4 & \multicolumn{1}{c|}{41.4±0.6} & 49.1±0.1 \\
\multicolumn{1}{c|}{SWAD}     & 59.4±0.5 & 57.4±0.6 & 54.1±0.4 & 58.8±0.4 & 54.8±0.1 & \multicolumn{1}{c|}{47.1±0.2} & 55.3±0.2 \\
\multicolumn{1}{c|}{RSC}      & 55.1±0.8 & 52.5±0.5 & 48.8±0.9 & 53.3±0.9 & 49.6±0.6 & \multicolumn{1}{c|}{42.1±0.8} & 50.2±0.6 \\
\multicolumn{1}{c|}{GroupDRO} & 53.4±1.0 & 51.8±1.1 & 47.9±0.8 & 53.2±0.3 & 50.5±0.5 & \multicolumn{1}{c|}{42.9±0.5} & 49.9±0.7 \\
\multicolumn{1}{c|}{Fishr}    & 54.4±0.4 & 52.1±0.8 & 47.9±0.5 & 51.8±0.2 & 48.5±0.8 & \multicolumn{1}{c|}{41.2±0.9} & 49.3±0.3 \\
\multicolumn{1}{c|}{CORAL}    & 52.5±1.6 & 51.1±0.6 & 47.2±0.9 & 53.0±1.5 & 48.7±1.5 & \multicolumn{1}{c|}{41.2±1.5} & 48.9±0.9 \\
\multicolumn{1}{c|}{MMD}      & 52.9±0.7 & 52.2±0.3 & 49.0±1.4 & 53.7±0.8 & 49.3±0.5 & \multicolumn{1}{c|}{42.2±0.5} & 49.9±0.5 \\
\multicolumn{1}{c|}{SagNet}   & 53.4±1.2 & 51.4±0.6 & 47.6±0.5 & 52.2±0.6 & 49.1±0.2 & \multicolumn{1}{c|}{40.8±0.6} & 49.1±0.1 \\
\multicolumn{1}{c|}{IRM}      & 53.3±2.3 & 51.0±0.9 & 46.2±1.1 & 52.1±0.9 & 48.8±0.7 & \multicolumn{1}{c|}{40.7±1.0} & 48.7±0.9 \\
\multicolumn{1}{c|}{Mixup}    & 62.7±0.1 & 60.5±0.6 & 57.5±1.0 & 62.4±0.5 & 58.5±0.3 & \multicolumn{1}{c|}{51.0±0.7} & 58.8±0.4 \\ \bottomrule
\end{tabular}%
}
\end{table*}

\end{document}